\documentclass{article}
\usepackage{iclr2025_conference,times}
\usepackage{soul}
\usepackage{amsmath,amsfonts,amssymb}
\usepackage{graphicx}
\usepackage{subcaption}
\usepackage{hyperref}
\usepackage{url}
\usepackage{wrapfig}
\usepackage{floatrow}
\usepackage{booktabs}
\usepackage[capitalize]{cleveref}
\usepackage{xspace}
\usepackage{multirow}
\usepackage{caption}

\crefname{table}{Tab.}{Tables}
\Crefname{table}{Table}{Tables}

\crefname{section}{Sec.}{Sections}
\crefname{appendix}{App.}{Appendices}

\iclrfinalcopy  

\definecolor{myblue}{HTML}{1971c2}
\definecolor{myorange}{HTML}{f08c00}
\definecolor{mygreen}{HTML}{4F7942}

\usepackage[normalem]{ulem}
\newcommand{\draftcomment}[3]{{\textcolor{#3}{[#1]#2}}}
\newcommand{\resolved}[1]{}
\newcommand{\roy}[1]{\draftcomment{#1}{\textsc{roy}}{red}}
\newcommand{\rs}[1]{\roy{\sout{#1}}}

\renewcommand{\rs}[1]{}

\newcommand{\knn}[0]{$k$-NN\xspace}
\newcommand{\wikitext}[0]{\textsc{WikiText-103}\xspace}
\newcommand{\pubmed}[0]{\textsc{PubMed}\xspace}
\newcommand{\mwiki}[0]{\textsc{Wiki40B}\xspace}

\title{From Tokens to Words: \\On the Inner Lexicon of LLMs
}
\author{Guy Kaplan, Matanel Oren, Yuval Reif, and Roy Schwartz \\
The Hebrew University of Jerusalem \\
\resizebox{0.95\textwidth}{!}{\texttt{\{guy.kaplan3,matanel.oren,yuval.reif,roy.schwartz1\}@mail.huji.ac.il}}
}

\begin{document}

\maketitle

\begin{abstract}
Natural language is composed of words, but modern large language models~(LLMs) process \textit{sub-words} as input. A natural question raised by this discrepancy is whether LLMs encode words internally, and if so how. We present evidence that LLMs engage in an intrinsic detokenization process, where sub-word sequences are combined into coherent whole-word representations at their last token. Our experiments show that this process primarily takes place within the early and middle layers of the model. We further demonstrate its robustness to arbitrary splits (e.g., ``cats'' to ``ca'' and ``ts''), typos, and importantly---to out-of-vocabulary words: when feeding the last token internal representations of such words to the model as input, it can 
``understand'' them as the complete word despite never seeing such representations as input during training. Our findings suggest that LLMs maintain a latent vocabulary beyond the tokenizer's scope. These insights provide a practical, finetuning-free application for expanding the vocabulary of pre-trained models. By enabling the addition of new vocabulary words, we reduce input length and inference iterations, which reduces both space and model latency, with little to no loss in model accuracy.\footnote{We release our code at \url{https://github.com/schwartz-lab-NLP/Tokens2Words}.}

\end{abstract}

\section{Introduction}

\begin{figure}
   \centering
   \includegraphics[width=0.8\columnwidth]{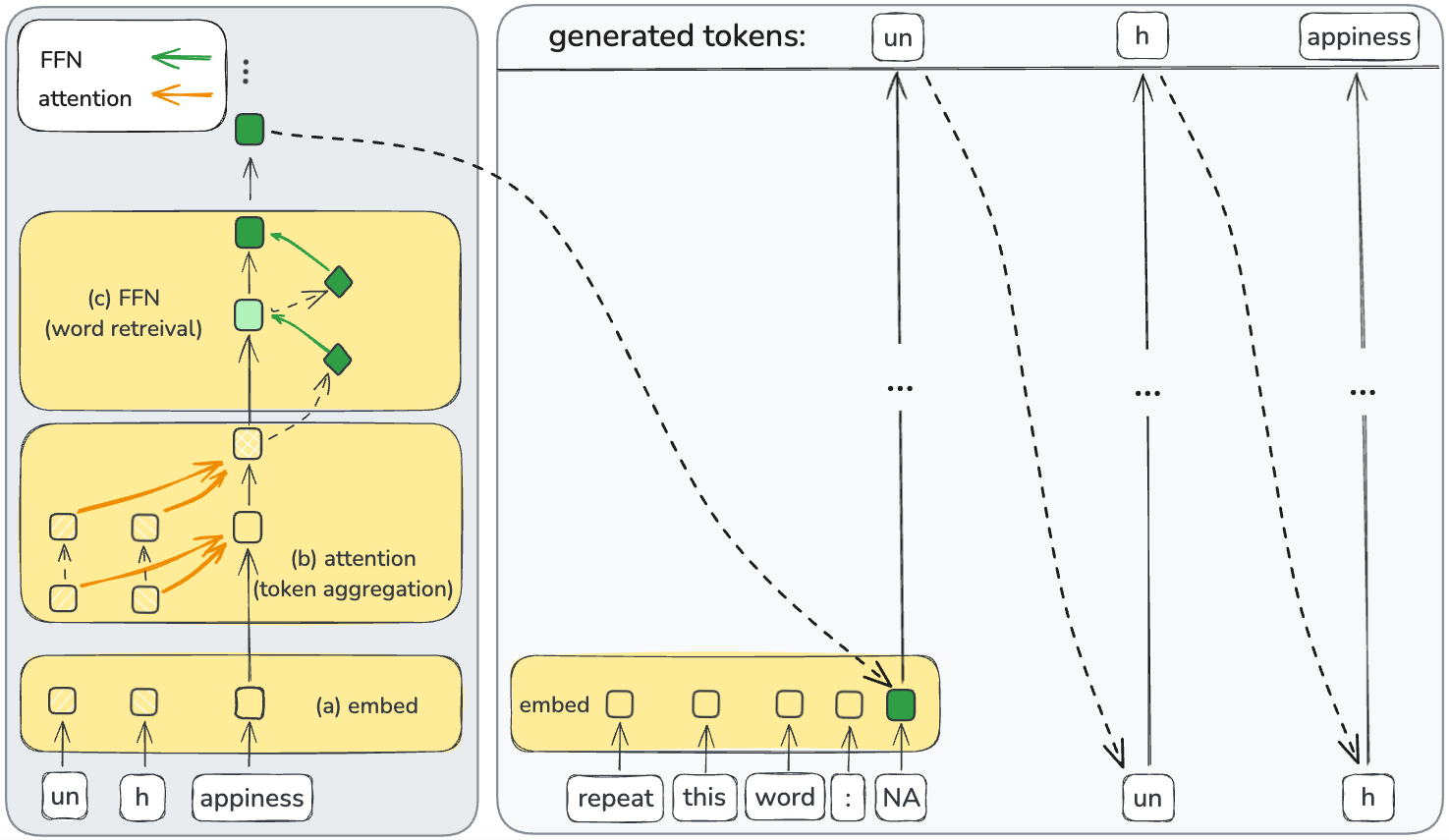}
   \caption{\textbf{Left}: The sub-word detokenization process in LLMs. From bottom to top:
   \textbf{(a) Tokenization and Embedding}: The input string is tokenized using a sub-word tokenizer (e.g., BPE) and converted into vector embeddings;
    \textbf{(b) Token Aggregation}:  The attention mechanism relays information from the word's previous tokens (``un'', ``h'') into its final sub-word representation (``appiness'');  % The attention mechanism is used to merge sub-word tokens into the final sub-word token
    \textbf{(c) Word Retrieval}: The model retrieves the full word representation from an implicit internal lexicon in its feedforward (FFN) layers. This representation is added to the residual stream, until it takes over the word's hidden representation. \\
    \textbf{Right}: When taking this hidden representation and patching it into another prompt, the model interprets it as the original word. In this example, the model is prompted to repeat the~(single vector) hidden representation of the word ``unhappiness'' (originally represented as three tokens), and is able to ``understand'' it by  regenerating the original three tokens.}
   \label{fig:DetokenizationFlow}
   % \vspace{-.25cm}
\end{figure}

Large language models (LLMs) rely heavily on tokenization methods such as byte-pair encoding~(BPE;~\citealp{sennrich-etal-2016-neural}). 
Such methods often split words into multiple tokens, 
potentially disrupting their morphological structure~\citep{batsuren2024evaluating}.\footnote{For instance, the word ``unhappiness'' might be tokenized into ``un,'' ``h,'' and ``appiness''~(see~\cref{fig:DetokenizationFlow})}  
Typos and other perturbations can also lead to large variations in the tokens that represent a word~\citep{kaushal-mahowald-2022-tokens}.
Nonetheless, LLMs exhibit a remarkable ability to recover word meaning~\citep{cao-etal-2023-unnatural}. %, implying the presence of  internal processes for word reconstruction.
This raises important questions 
about how models internally compose meaningful word representations from tokens, 
a process referred to as \textit{detokenization}~\citep{elhage2022solu, gurnee2023finding}.

In this work, we seek to understand the detokenization mechanism in LLMs. We consider two cases: words that are not part of the model's BPE vocabulary, and are thus split into multiple sub-words; and single-token, in-vocabulary words that we artificially split into multiple tokens. 
In both cases, our experiments indicate a word-level detokenization process in LLMs, which occurs mainly in the early to middle layers. 
Our results hint that language models hold a \emph{latent vocabulary} or \emph{inner lexicon}, which they access to identify words from token sequences.\footnote{This process might resemble the \textsl{mental lexicon} in humans~\citep{aitchison2012words, marslen1994morphology}.}

We begin by examining if internal representations of token sequences reflect whether or not a sequence of tokens comprises a word~(\cref{sec:word_vs_gibberish}).
We probe the model's hidden representations \citep{Probing} of both multi-token real English words and gibberish \emph{nonwords}~\citep{Frisch2000PerceptionOW}. We observe that the representations of words and nonwords substantially diverge in middle layers---a simple k-nearest neighbors classifier achieves a 89\% accuracy on the task of discriminating between the two groups. 
Overall, these results suggest that models hold a concept of recognized words.

We next explore the mechanism through which models reconstruct cohesive word representations from sub-word tokens~(\cref{sec:when_detokenization}). To do so, we use techniques that interpret a token's hidden states and decode them into natural language \citep{belrose2023elicitinglatentpredictionstransformers, Ghandeharioun2024PatchscopesAU}. 
We find that for both multi-token and single-token words, in most cases, the last token can be decoded as the full word after being processed by 3--5 layers,
with some words requiring up to 15 layers. 
Interestingly, 23\% of the multi-token words are never successfully decoded from their last token's hidden states, hinting this inner lexicon does not cover all words.

We then turn to explore how models assemble these full word representations~(\cref{sec:how_detokenization}). 
We first interpret the feedforward network (FFN) updates in the vocabulary space \citep{geva-etal-2021-transformer, todd2024function}. We show that in 85\% of the evaluated words, an FFN update promoting the full word's concept is written to the last token's residual stream  \citep{geva-etal-2022-transformer, merullo-etal-2024-language}. Importantly, this happens in the layers \textit{before} the full word's representation emerges, indicating that models retrieve the reconstructed word representation from their FFN weights. 
Second, we compare attention patterns between standard single-token words to how multi-token words attend to their preceding sub-word tokens. We find that the last token in multi-token words attends significantly \emph{more} to its previous tokens~(which are sub-words of the same word) in layers 1--2 compared to single-token words~(where these previous tokens represent other words), and significantly \emph{less} in the following layers. This suggests that models initially aggregate information from previous tokens, and later almost ignore them. See~\cref{fig:DetokenizationFlow} for an overview.

Our findings have concrete applications~(\cref{sec:vocab_expansion}); they allow us to expand the model input and output BPE vocabulary with the fused representation of multi-token words found in our experiments. Importantly, this expansion requires no update to model parameters. Applying our approach to multi-token words found in English Wikipedia data~\citep{wikitext}, we find that the model successfully uses the new vocabulary entries both as inputs and during generation: its language model performance is maintained, and even slightly improves. This demonstrates the potential to dramatically reduce both input and output sequence length, and inference costs accordingly, especially in languages where the ratio of tokens per word is high~\citep{ahia-etal-2023-languages, petrov2024language}.

Overall, our results establish that word-level detokenization is a core process in LLMs, and provide evidence of how it unfolds across model layers. Beyond improving our understanding of the internal mechanisms driving LLMs, our work lays the foundation for practical applications, particularly in  optimizing token management and reducing computational costs.

\section{Related Work}

\paragraph{Tokenization}
Sub-word tokenization algorithms~\citep{wu2016wordpiece, kudo-2018-subword} are the standard for pre-processing text in modern LLMs. The most widely used method is byte-pair encoding~(BPE;~\citealp{sennrich-etal-2016-neural}), which keeps frequent words intact and splits rare ones into multiple sub-words. 
Recent studies proposed ways to improve tokenization to consider word structure~\citep{provilkov-etal-2020-bpe, hofmann-etal-2022-embarrassingly, yehezkel-pinter-2023-incorporating, bpe_knockout} or analyze how tokenization affects model performance~\citep{bostrom-durrett-2020-byte,church2020emerging,klein2020getting,zouhar-etal-2023-tokenization,schmidt2024tokenization}. To the best of our knowledge, we are the first to thoroughly investigate how LLMs internally reconstruct word representations.

\paragraph{Detokenization and stages of inference} 
Early LLM layers have been shown to integrate local context and map raw token embeddings into representations of concepts or entities---a process called \emph{detokenization}. However, such observations were based on specific case studies~\citep{elhage2022solu,lad2024remarkablerobustnessllmsstages}. %Prior work probed transformer models to explore how predictions are constructed over the course of the model,
More generally, recent work showed early layers provide local syntactic information~\citep{BERT_Rediscovers,vulic-etal-2020-probing, durrani-etal-2020-analyzing,sajjad-etal-2022-analyzing} or focus on extracting information from previous tokens~\citep{artzy-schwartz-2024-attend}. 
\cite{ferrando2024information} analyzed token attributions in LLMs, observing attention heads that promote sub-word merging~\citep{correia-etal-2019-adaptively}.
Our work focuses on word-level detokenization, and goes beyond previous efforts to provide an in-depth analysis of how word representations are assembled from multiple tokens.

\paragraph{Interpreting the residual stream} 
Recent methods for interpreting the intermediate states of LLMs draw on a \emph{residual stream} perspective: the hidden state acts as an information stream along the layers, from which information is read at each layer, and new information is added through residual connections \citep{elhage2022solu}. Thus, hidden states at any layer can be projected into the model's vocabulary space, treating the hidden state as if it were the output of the last layer~\citep{nostalgebraist2020logitlens,dar-etal-2023-analyzing,belrose2023elicitinglatentpredictionstransformers,yom-din-etal-2024-jump}.
Similarly, \cite{Ghandeharioun2024PatchscopesAU} proposed to decode information from hidden representations into natural language, by patching \citep{zhang2024towards} it into a prompt that encourages the model to \emph{verbalize} the encoded information.
We use both approaches to inspect how token representations evolve across layers.

\paragraph{LLM memories}
The idea of an \emph{inner lexicon} aligns with recent work showing feedforward networks~(FFN) layers in transformers act as key-value memories that encode factual and linguistic knowledge~\citep{geva-etal-2021-transformer, meng2022locating, dai-etal-2022-knowledge}. Particularly, FFNs were shown to enrich entity tokens with associated information~\citep{ meng2022locating,geva-etal-2023-dissecting} and promote relevant concepts in vocabulary space to build up predictions~\citep{geva-etal-2022-transformer}. Our work expands on these findings, showing that word representations are pulled from FFN layers before emerging in the hidden state of the word's last token.

\paragraph{Inner lexicon structure}
Inspired by studies on how concepts are encoded in LLMs~\citep{park2024geometry,LRH} and gradually promoted throughout their layers~\citep{geva-etal-2022-transformer,merullo-etal-2024-language}, we consider the \textit{inner lexicon}  a ``soft'' lexicon, which~(a) combines multiple vectors to form word representations~(rather than a key-value dictionary); and~(b) is not unique, i.e., a word might be stored and retrieved in more than one layer. 
Concurrently,~\cite{feucht2024tokenerasurefootprintimplicit} present evidence for an implicit vocabulary in LLMs, showing that models ``forget'' preceding tokens in multi-token words or multi-word entities, but remember previous tokens when processing single-token words. 

\section{A Motivating Observation: LLMs Can Tell Words From Non-Words}
\label{sec:word_vs_gibberish}

One of our key hypotheses in this work is that LLMs hold an internal lexicon of words, which is different from the BPE lexicon. 
We begin by asking whether LLMs, when processing a sequence of tokens, capture some notion of whether or not this sequence forms a word.

\begin{figure}[t]
    \centering
    \begin{subfigure}{0.48\textwidth}
        \centering
        \includegraphics[width=1\textwidth]{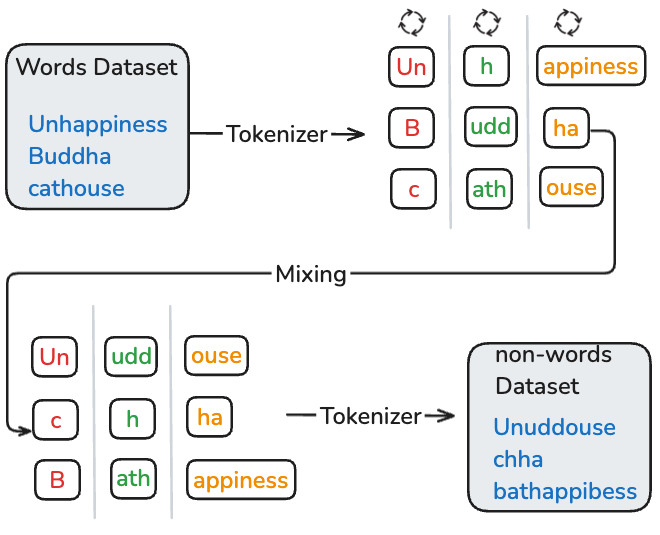}
        \caption{Nonword creation process}
        \label{fig:dataset_creation}
    \end{subfigure}
    % \hill
    \begin{subfigure}{0.49\textwidth}
        \centering
        \includegraphics[clip,width=\textwidth]{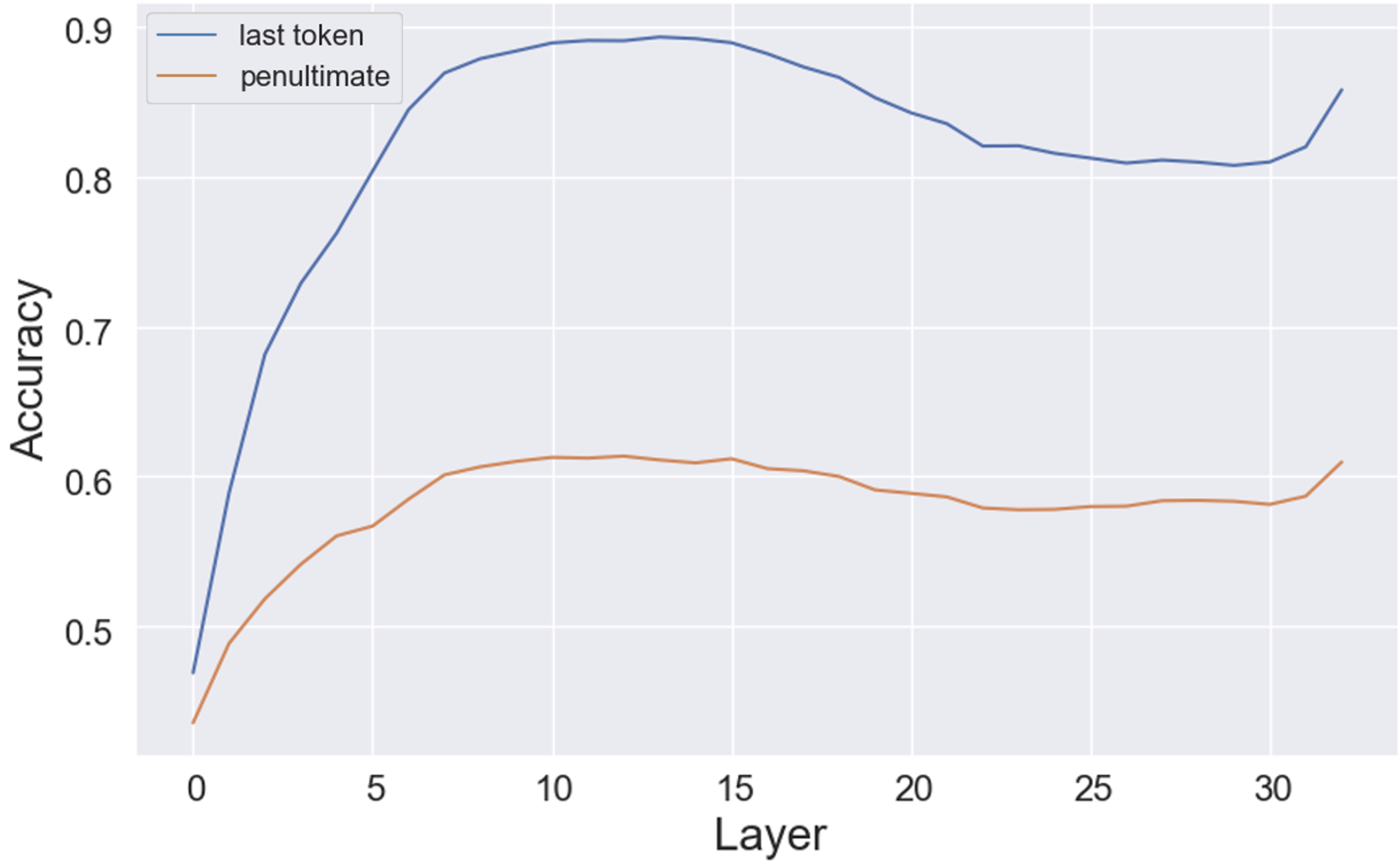}
        \caption{Classification results}
        \label{fig:validation_accuracy_nonwords}
    \end{subfigure}
    \caption{Our word vs.~nonword probing classification experiments. (\ref{fig:dataset_creation}) \textbf{Dataset creation process}. Top: words from the Gutenberg corpus are tokenized using the Llama2 tokenizer. Bottom: nonwords are generated by shuffling 
    tokens while maintaining their positions within the word; (\ref{fig:validation_accuracy_nonwords})~\textbf{Classification results of words vs.~nonwords}. Using the {\textcolor{myblue}{last}} token shows strong results (reaching up to 90\% accuracy), which sharply rise in early layers~(3--7), peak 
    in the middle~(13), and decrease in later layers~(20--32). 
    Using the  {\textcolor{myorange}{penultimate}} token shows substantially lower scores, suggesting that the high classification accuracy is specifically tied to the presence of a complete word rather than token co-occurrence patterns.}
    \label{fig:dataset_and_prediction}
\end{figure}

To address this question, we construct a balanced dataset containing two groups: one with real English words, and another with artificially generated, meaningless \emph{nonwords}~\citep{Frisch2000PerceptionOW}. Both groups are tokenized using the Llama2 tokenizer \citep{touvron2023llama2openfoundation}. 
The word dataset consists of 10,000 distinct words sampled from the Gutenberg corpus \citep{gerlach2018standardizedprojectgutenbergcorpus}, with 53\% of the words containing two tokens, 37.3\% containing three tokens, and the rest four tokens. 
We generate the nonwords by shuffling tokens from the word dataset, ensuring that the prefix and suffix positions align with the original tokens' positional probabilities.
For example, the token “ing”, extracted from the final position of real words in the dataset (e.g., “directing” tokenized as ``direct'', ``ing''), is retained as a suffix in the nonwords dataset.
This process ensures that nonword tokens maintain position distribution properties similar to word tokens, preserving natural positional patterns and mitigating potential distributional biases~(\cref{fig:dataset_creation}).
We next apply a $k$-nearest neighbors~(\knn) probing classifier ($k=4$, using Euclidean distance) on the hidden states of the last tokens of both words and nonwords, for each layer of the Llama2-7B model. 
The training set consists of 80\% of the dataset, and the remaining 20\% are used for evaluation.

% \begin{wrapfigure}{r}{0.5\textwidth}

Our results~({\cref{fig:validation_accuracy_nonwords}, {\textcolor{myblue}{blue}}) reveal a three-stage pattern in the model’s representation  of word and nonword token sequences. 
In the model's first few layers, representations from both groups are relatively indistinguishable and accuracy is close to chance level. Then, from layers 2 to 6, a clear distinction between the two groups emerges, until the representations are almost completely separate in middle layers, between layers 6 and 20. At this point, the probe achieves a stable, high accuracy, peaking at 89\% on layer 13. 
Finally, accuracy slightly drops after layer 20.\footnote{We perform a similar experiment using a dataset of morphologically plausible nonwords (ARC Nonword Database; \citealt{rastle2002arc}), which shows a similar trend. See \cref{sec:appendix_words_vs_nonwords}.}

Our results indicate the LLMs can distinguish between words and nonwords. But this distinction can be attributed to the distributional properties of words: it might be the case that models are recognizing common sequences of tokens, rather than identifying whole words. To test this hypothesis,  we repeat the same experiment, this time using the \textit{penultimate} token representation, for words three tokens or longer. By definition, such words also as frequently co-occur with the initial sub-word tokens in their words as the final tokens.  Our results~(\cref{fig:validation_accuracy_nonwords}, {\textcolor{myorange}{orange line}}) show that the probe only reaches 61\% classification accuracy, indicating that the high classification accuracy does not stem from token co-occurrence, but is tied to the presence of a whole word. See \cref{sec:appendix_words_vs_nonwords} for an analysis of misclassified words and nonwords.

Overall, our results show that language models internally represent words and nonwords differently. This distinction is gradually developed in the model's early layers and maintained throughout its middle layers.
These findings support the hypothesis that the model performs a detokenization process and suggests where this process occurs.
Building on these results, we next investigate how sub-word tokens are combined into word representations across model layers, and explore the internal mechanisms that facilitate this transformation.

\section{Extracting Word Identity From LLM Hidden States}
\label{sec:when_detokenization}

We have so far observed that LLMs can differentiate between words and nonwords, suggesting an internal detokenization process specific to word composition.
We next dig into this process, by asking whether we can directly extract word identity from the hidden states of sub-word tokens. We start by considering single-token words~(\cref{sec:Single-Token_Words_Exp}), and then move on to multi-token words~(\cref{sec:Multi-Token_Words_Exp}).

\subsection{Single-Token Words}
\label{sec:Single-Token_Words_Exp}

We first consider in-vocabulary words, which are mapped to single tokens by the tokenizer. Naively, such words don't tell us much about detokenization, as they are represented using only one token. To address this, we artificially split them into multiple sub-word tokens. For example, we take the single-token word ``cats'' and split it into two tokens: ``ca'' and ''ts''. We hypothesize that if the model performs detokenization, it will represent the last token of the word (``ts'') similarly to the original word token~(``cats''). We iterate the \wikitext dataset~\citep{wikitext} and randomly split each single-token word longer than three characters into 2--5 sub-words tokens. We then feed the model the new sequence of tokens preceded by the last 100 tokens that came before each split word in the original text as context.\footnote{We observe similar trends, in this and in other experiments in this section, when passing the split words \emph{without} their context, albeit with slightly lower rates of retrieval.}

To measure the similarity between the representation of the final token and original word, we apply the logit lens method~\citep{nostalgebraist2020logitlens}, an interpretability method that maps a hidden representation of a given token to the word whose vector in the output unembedding matrix is most closely aligned with the hidden representation.
We note that this technique is typically used to inspect model predictions of the \emph{next} token in intermediate layers.  As we aim to identify how LLMs represent the \emph{current} word, we use the \textit{input} embeddings matrix rather than the unembedding one.\footnote{We also compute a second measure, cosine similarity, to compare between the hidden representations and input embeddings. We find  similar results to those obtained using logit lens. For details, see~\cref{sec:appendix_logit_lens_vs_cosine_similarity}.}

We study four LLMs: Llama2-7B, Llama3-8B~\citep{dubey2024llama3herdmodels}, Mistral-7B~\citep{jiang2023mistral7b}, and Yi-6B~\citep{ai2024yiopenfoundationmodels}. For each layer, we report the rate of retrieval---the proportion of words for which the closest vector in the vocabulary space is the original (single token) word. 

Our results for the Llama2-7B are shown in~\cref{fig:rate_single_tokens}.\footnote{Results for the other models (\cref{sec:appendix_all_models_word_retrieval}) show a similar trend.} Starting at layer 8, the hidden state of the final token is mapped to the original word with high accuracy, which peaks at more than 80\% in layer 15. Interestingly, this accuracy then starts to decline, possibly because the model transitions into representing the prediction of the next token. We also find that 93.2\% of all split words are correctly mapped to the original word in \textit{at least one} model layer~(see~\cref{sec:appendix_cummulative_word_retrieval}).

Our results indicate that LLMs perform a detokenization process in cases where a single-token word is split into multiple tokens. This process assigns the final token with a hidden representation similar to that of the full, single-token word. But \textbf{how robust is this process}? To address this question,
we consider a different way of splitting a word into sub-tokens---adding typos. For each word, we randomly flip two adjacent characters, delete a character, or insert a new character (see~\cref{sec:appendix_introducing_typos} for more details). Much like our previous experiment, this process splits the word into multiple tokens, as corrupted words are rarely found in BPE lexicons. We repeat the same experiment as above, aiming to study whether models map the last token to the original~(correctly-spelled) word. 

Our results for Llama2-7B~(\cref{fig:rate_single_tokens}) show a similar trend, though less pronounced: the model correctly matches about 40\% of corrupted words with the original, single-token, uncorrupted word.\footnote{Here, 66\% of the words are correctly mapped in at least one layer.}  We also note that the trend across layers persists, with accuracy peaking around layer 15.

\begin{figure}[t]
    \centering
    \begin{subfigure}{0.48\textwidth}
        \centering
        \caption{Rate of retrieval for single-token words}
        \includegraphics[width=\textwidth]{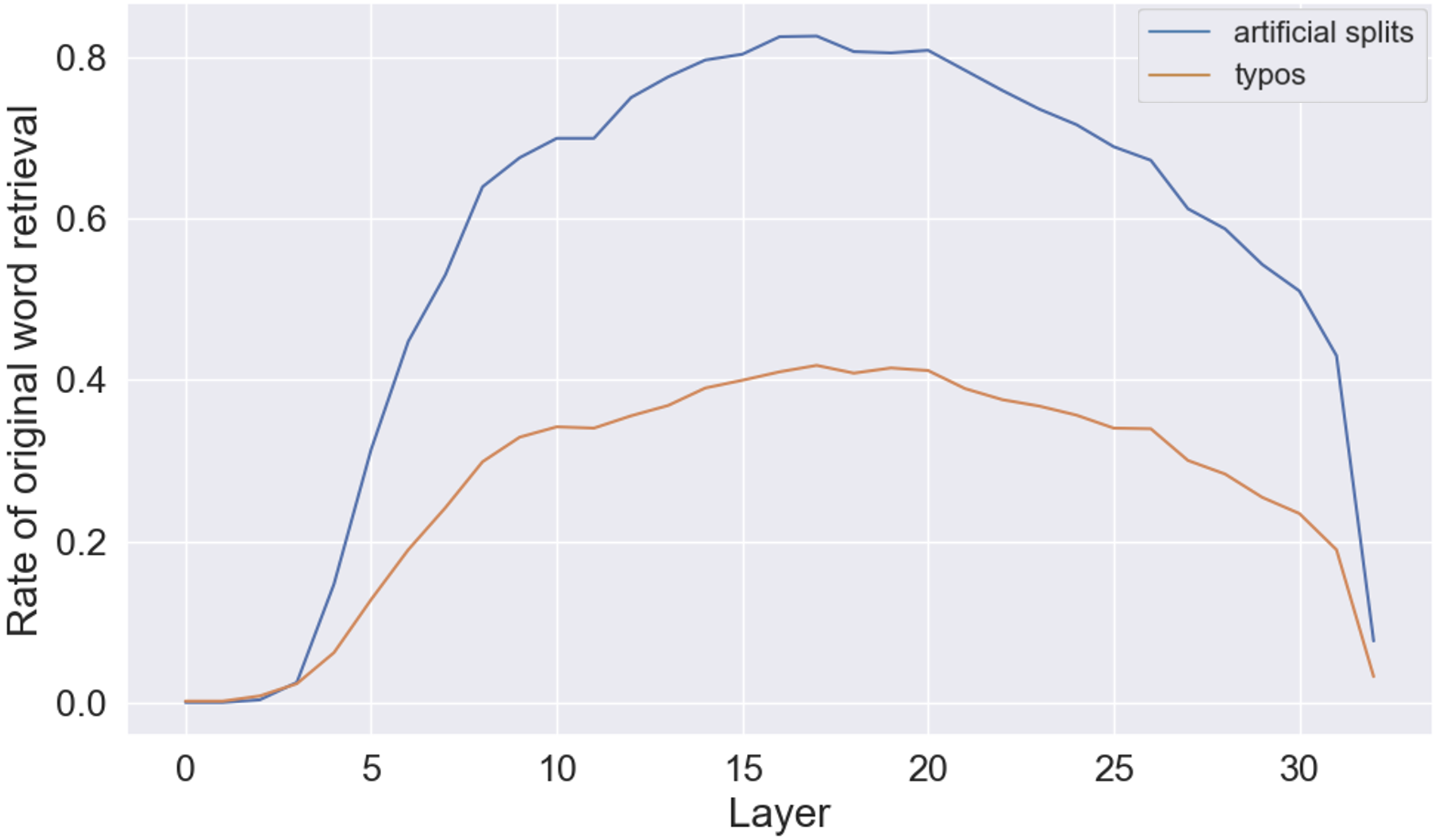}
        \label{fig:rate_single_tokens}
    \end{subfigure}
    \hfill
    \begin{subfigure}{0.48\textwidth}
        \centering
        \caption{Rate of retrieval for multi-tokens words}
        \includegraphics[width=\textwidth]{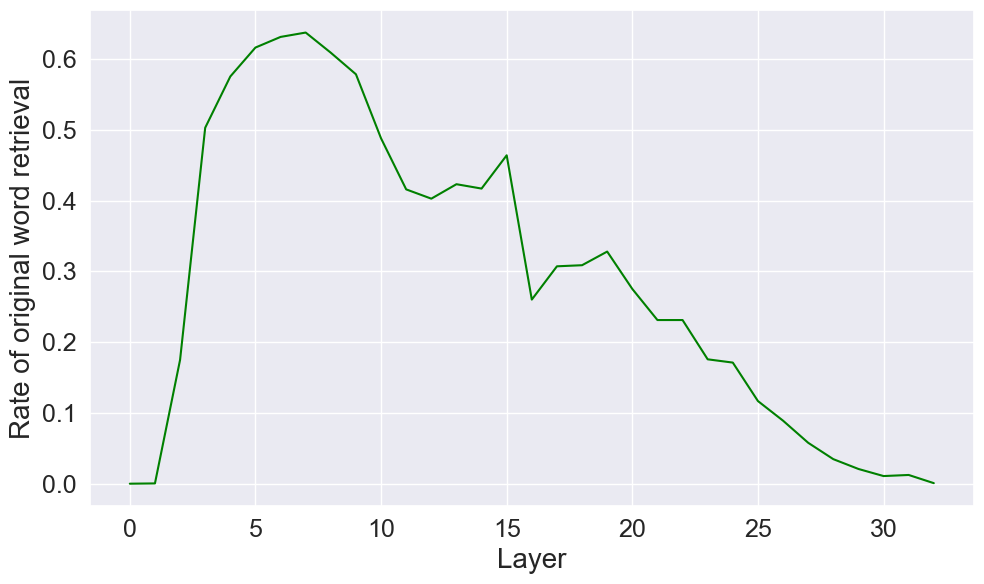}
        \label{fig:rate_multi_tokens}
    \end{subfigure}
    \caption{~(\ref{fig:rate_single_tokens})
    Logit lens rate of retrieval of single-token words artificially split ({\textcolor{myblue}{blue}} line) and split due to typos ({\textcolor{myorange}{orange}}).
    In both cases, we see an increasing rate of retrieval after the 4th layer, peaking in the middle (16--17 layers) and then dropping;~(\ref{fig:rate_multi_tokens})
    Patchscopes rate of retrieval for multi-token words. In this case results peak in the 5th layer and then start to decline.
    }
\end{figure}

\subsection{Multi-Token Words}\label{sec:Multi-Token_Words_Exp}
We have seen that LLMs are able to detokenize single-token words that are artificially broken into sub-words back into their correct (single-token) word. This indicates that these words are part of the model's inner lexicon. But what about multi-token words? By definition, these words do not have a single embedding vector in the BPE vocabulary, so we cannot compare them to any existing vector, and thus cannot use logit lens.

Instead, we use the Patchscopes technique~\citep{Ghandeharioun2024PatchscopesAU}, which interprets the hidden representation of a model using the model's language abilities. 
In particular, we feed the model with the prompt ``Repeat this word twice: 1) $X$ 2)'', where $X$ is the hidden representation of the last word token. Our hypothesis is that if the multi-token word is found in the model's internal lexicon, then it will ``understand'' its representation in the input layer as well, and successfully repeat it by generating all tokens of the original word~(see~\cref{fig:DetokenizationFlow}, right).

We evaluate the model by computing the rate of retrieval---the proportion of times it generates the correct~(multi-token) word.
We use the same models and dataset as in the previous experiments, though in this case we do not further split or add typos, as the words are already split into multiple tokens. Particularly, we feed each multi-token word (along with its 100-tokens context, as before) to the model, and extract the last token's representation from each layer. We then use Patchscopes' prompt in a new run to check if the model regenerates the original multi-token word.
}

Our Llama2-7B results~(\cref{fig:rate_multi_tokens}) show a striking trend: when feeding the model with vectors from layers 5--7, it is able to repeat the word in 64\% of the cases, despite never seeing this vector as input before.\footnote{Overall, 77.4\% of the words are repeated correctly in at least one layer.} As to the different layers, we observe a similar trend to the single-token results: retrieval rates increase rapidly at some point (here earlier than in previous experiments, around layer 3-4), reach a peak at layer 7, and then start to decline. 
Combined, these results suggest that the model
treats multi-tokens words as if they were in the vocabulary but split into sub-word tokens, indicating a latent vocabulary that expands beyond the tokenizer's limitations. Interestingly, we observe that 22.6\% of the multi-token words are never successfully decoded from any internal layer, hinting that they might not be represented in the model's inner lexicon.

\section{How Does Detokenization Happen?}
\label{sec:how_detokenization}

We have so far shown that LLMs perform a detokenization step: at some point, the hidden representation of the final token of a given word becomes strikingly similar to the (single-tokenx) vector of that word. This process is robust to artificial splits of single token words, to splits due to typos, and even to multi-token words, which the model can still recognize at the input layer, despite never having seen them there during training.

We next turn to ask \textbf{how} does the model reconstruct full word 
representations from sub-word tokens? 
We aim to understand the dynamics of this process by analyzing the main transformer components: feedforward network layers (FFN) and the attention mechanism.

\subsection{Word Retrieval Using the Feedforward Mechanism}

FFN layers have been shown to serve as key-value tables for storing memory~\citep{geva-etal-2022-transformer, geva-etal-2021-transformer,meng2022locating}. We hypothesize that this memory might be used to store the inner lexicon as well.
Particularly, we suggest that the model uses the FFNs to refine the representation of the final token, enabling it to retrieve the original word.

\begin{figure}[t]
    \centering
    \begin{subfigure}{0.48\textwidth}
        \centering
        \caption{Word retrieval per layer}
        \includegraphics[width=\textwidth]{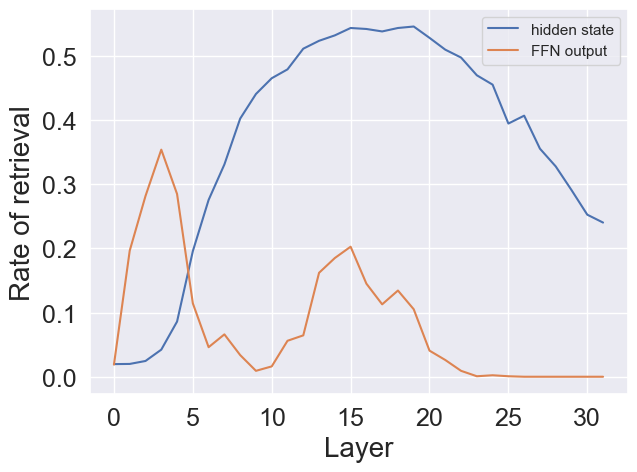}
        \label{fig:word_retreival_ffn}
    \end{subfigure}
    \hfill
    \begin{subfigure}{0.48\textwidth}
        \centering
        \caption{Cumulative word retrieval per layer}
        \includegraphics[width=\textwidth]{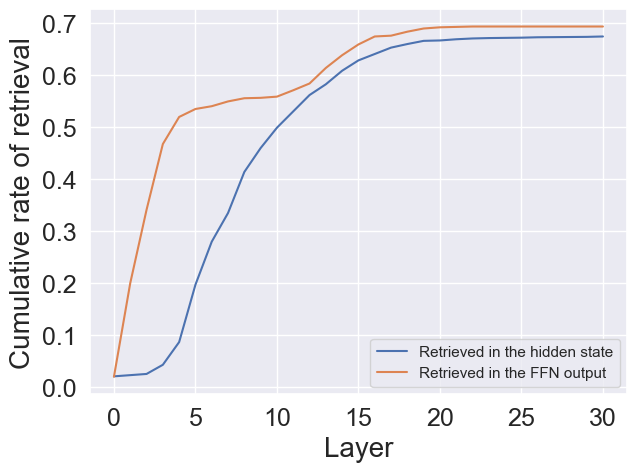}
        \label{fig:word_retreival_ffn_cumulative}
    \end{subfigure}
    \caption{
    Word retrieval in FFN layers vs.~in the hidden states. (4a) Retrieval rates across layers. The FFN values peak before word retrieval begins in the hidden layers. (4b) Its cumulative version, showing word retrieval occurs earlier and more frequently in FFNs than in the hidden representation.
    }
\end{figure}

To test this hypothesis, we repeat the typos experiment of single-token words~(\cref{sec:Single-Token_Words_Exp}),  
but this time applying logit lens to the \textit{output of the FFN layers}, instead of the hidden representation. \Cref{fig:word_retreival_ffn} shows the retrieval rate of the original word. 
We observe that the retrieval of the full word concepts from the FFN layers occurs a few layers earlier compared to the hidden state, which indicates that they help refine it. Nonetheless, the rate of retrieval appears to be lower in FFNs, which suggests that other factors might come into play in building the hidden representation of words. However, in~\cref{fig:word_retreival_ffn_cumulative} we plot the cumulative retrieval rate, i.e., for each layer, whether the word is identified in \textit{any} layer so far. Our results indicate a different conclusion:
the FFN update vectors match the (single-token) input representation of the word  70\% of the time in any layer, particularly both earlier and more frequently compared to the hidden representation. This hints that FFNs indeed play a substantial role in building the internal word representations in LLMs.

We further investigate the role FFNs play in detokenization through ablation experiments on FFN updates in Llama2-7B. Specifically, we test whether these updates are necessary for the word representation to emerge in the residual stream.
To do so, we artificially split single-token words in a similar setting to \cref{sec:Single-Token_Words_Exp}, this time focusing on derived words, formed by adding a suffix to a root word.\footnote{We examine words with three common suffixes: ``ing'', ``ion'', and ``est''.} We split these back to two parts, for example, we divide ``eating'' to ``eat'' and ``ing'', so that processing the suffix token is essential to reconstructing the word correctly.
Using logit lens, we examine the FFN updates to the suffix token's residual stream, and selectively ablate those associated with the original single-token word ($\sim$5\% of layers). As a control, for each word, we ablate an equal number of random FFN updates. Our results (\cref{fig:ablated_ffn_vs_regular} in \cref{sec:appendix_ffn_ablation}) show that removing the updates carrying full-word representations dramatically reduces retrieval rates—from 85\% without ablation to just 18\%. In contrast, ablating random FFN updates has little to no effect.

This indicates that FFN updates are essential for a detokenized representation to emerge in the final token. But does this affect the model’s ability to "understand" the word, particularly when predicting the next token? To investigate this, we evaluate the model's ability to retrieve the capital cities of countries, using the prompt ``The capital of [COUNTRY] is \_\_\_\_\_\_''. We use single-token country names artificially split into two tokens, and follow the previous ablation protocol.
We observe notable patterns: randomly ablating FFN updates reduces performance from 88\% to 74\%,\footnote{We note that baseline performance when passing country names without artificial splits is 95\%.} likely due to disrupting the model's factual recall mechanism~\citep{meng2022locating}. However, specifically canceling updates detected as country’s name (similarly occurring in only $\sim$5\% of layers) causes a sharp drop to 41\%. Overall, our results suggest that FFNs are central to detokenization, and actively reconstruct full-word meanings in LLMs.

\subsection{Token Aggregation}\label{sec:token_aggregation}

Our results suggest that LLMs use the FFN layers to store and retrieve word representations, which are accessed in the final token to reconstruct the complete word.
However, this role of the FFN layers only begins to emerge around layers 3--4. \textbf{What happens earlier}?
We build on previous work showing that early layers primarily integrate information from nearby tokens to compose entities~\citep{lad2024remarkablerobustnessllmsstages}, and hypothesize that the model starts by aggregating information from the previous sub-word tokens.

\begin{wrapfigure}{r}{0.48\textwidth}
\vspace{-12pt}
  \centering        
  \includegraphics[width=\textwidth]{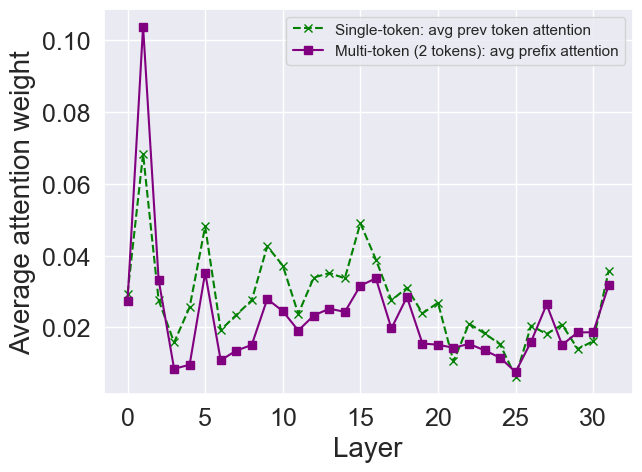}
        \caption{Attention weights for 2-token words: Early peaks (layers 2--3) show high 
        attention values from the second sub-word token to the first, but these values decline rapidly. 
        Attention from single-token words to their previous token shows a similar trend, though with 
        substantially lower values at first, which become higher later.}
        \label{fig:Token_Aggregation_experiment}
        
\vspace{-10pt}
\end{wrapfigure}

To test this, we extract all two-token words in a subset of 1,000 
\wikitext documents~(a total of 5,571 words), and feed them, along with their context, to Llama2-7B. We then measure the average attention weights of the final token to the prefix token in each layer.
As a control, we also measure the average attention weights assigned by single-token words to their preceding token. 

Our results (\cref{fig:Token_Aggregation_experiment}) support previous findings~\citep{lad2024remarkablerobustnessllmsstages}---the attention to previous tokens is high
in the first 2--3 layers, but then declines sharply~(by up to $\sim$90\%).\footnote{Experiments with 3- and 4-token words show a similar trend, see~\cref{sec:appendix_token_aggregation}.}
For single-token words, we observe a similar attention pattern to the previous token in the first layers, but importantly---the initial peak is significantly lower than in multi-token words.\footnote{The diverging patterns between single and multi-token words are statistically significant; see~\cref{sec:appendix_token_aggregation}.} Still, in later layers, the attention weights of single-token words are in turn \emph{higher} than for multi-token words. These results suggest LLMs strongly attend to the preceding sub-word tokens of multi-token words \emph{at first}, but then largely ignore them.

Altogether, our results suggest that LLMs perform a detokenization process by first aggregating information from the prefix tokens into the final token's hidden representation, and then refining the representation of the final token using the FFN layers to retrieve the full word's concept representation. 
This two-stage process of token aggregation and concept retrieval provides insight into the mechanisms LLMs use to handle sub-word tokens and reconstruct word-level representations.

\section{Expanding LLM Vocabulary Without Finetuning}
\label{sec:vocab_expansion}

We have shown that language models internally fuse multi-token words into a single-token representation, and that they can further ``read'' these representations as inputs, and decode the original multi-token words. This raises the question: can models use these fused representations instead of the original multi-token inputs---to encode input prompts using less tokens and reduce computation? Similarly, models were shown to implicitly predict several future tokens in a single hidden state without being explicitly trained to do so~\citep{pal-etal-2023-future}; can we leverage these representations to enable models to predict multi-token words in a single inference step? 

Motivated by our findings, we explore whether we can expand the model's vocabulary with new input and output embeddings for originally multi-token words, without any updates to model parameters.\footnote{We do add new entries to the model's embedding and unembedding matrices, but these are intrinsically required to expand the vocabulary and represent the new vocabulary words.}
This goal is of practical importance, especially for low-resource languages and domains:  even tokenizers built for multilingual support often produce substantially longer token sequences for non-English languages---up to 13 times longer than equivalent English texts---impacting inference cost and speed~\citep{ahia-etal-2023-languages,sengupta2023jais, petrov2024language}.
Still, prior attempts to expand tokenizer vocabulary post-hoc are limited in number, and require substantial additional training~\citep{kim2024efficient,zhao2024llama}. In contrast, we propose to detect words the model ``knows'' and successfully detokenizes to a single vector, and use this to obtain new token embeddings.

\begin{figure}[h]
    \centering
    \includegraphics[width=0.9\textwidth]
    {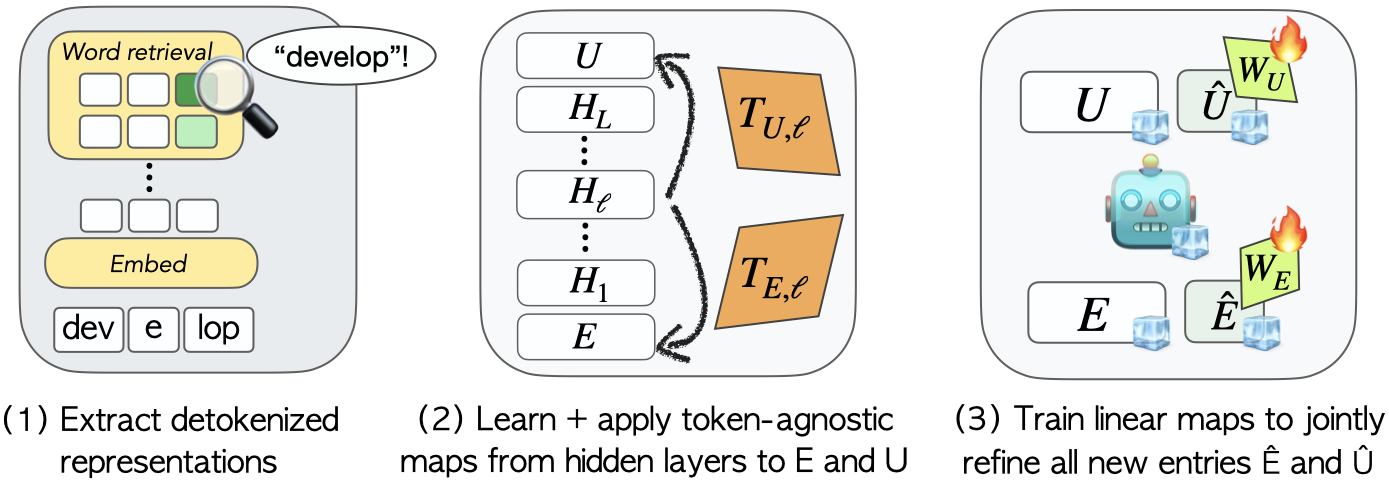}
    \caption{
    Our 3-step method to expand LLM vocabulary without updates to core model parameters.
    }
\label{fig:vocab_expansion_method}
\end{figure}

Our framework for vocabulary expansion follows a 3-step process (\cref{fig:vocab_expansion_method}). Given a multi-token word $w$ that we would like to add to the vocabulary, and the model's original input embedding and output unembedding matrices $E$ and $U$ with hidden dimension $d$, we (1) extract a detokenized, single-token representation $r$ for $w$ using a similar approach to~\cref{sec:Multi-Token_Words_Exp}: pass $w_i$ as input to the model and apply \textsc{Patchscopes}~\citep{Ghandeharioun2024PatchscopesAU} with prompt $P$\footnote{For $P$, we use the template ``x x x x'' where x is the patched representation of the new word.} to the last token's hidden states at all layers. We then identify $\ell$, the earliest layer at which the hidden state is successfully decoded into the full word, and set $r$ as that hidden state. Next, we (2.1) learn a set of linear maps $T_{\ell, E}$ and $T_{\ell, U}$ to project hidden states from layer $\ell$ of the model to the embedding and unembedding spaces, based only on the \emph{existing} in-vocabulary tokens.\footnote{We learn $T_{\ell,E}$ and $T_{\ell,U}$ by fitting an orthogonal procrustes transformation from the layer $\ell$ hidden states of all in-vocabulary tokens (when passed as inputs on their own), to their corresponding embeddings or unembeddings (see~\cref{sec:appending_learning_procrustes_maps} for details). Importantly, these projections only rely on the model's existing vocabulary, and do not depend on which multi-token words we choose to add to the vocabulary.} Then, (2.2) for each detokenized representation $r$ taken from layer $\ell$, we apply $T_{\ell,E}$ and $T_{\ell,U}$ to obtain the initial entries in the embedding and unembedding matrices to represent the word, $\hat{e}$ and $\hat{u}$. Finally, after computing the initial entries for all new words, we (3)~refine the new representations to obtain the final entries: we initialize two $d\times d$ all-zeros matrices $W_E$ and $W_U$, and set the new entries as $e = \hat{e} + W_E \hat{e}$ and $u = \hat{u} + W_U \hat{u}$. We train the refinement matrices $W_E$ and $W_U$ jointly in a short continued pretraining run, while keeping all other parameters frozen.  
Finally, we compute the final $e$ and $u$, and use these to represent the new word in the expanded vocabulary. Importantly, if a word is never successfully decoded from the hidden states in any of the layers in step (1), we assume it is not found in the model's inner lexicon, and therefore do not add it to the vocabulary.

We apply our approach to  \textsc{Llama2-7B} and experiment with three datasets: \wikitext~\citep{wikitext}, abstracts of biomedical articles from \pubmed~\citep{pubmed}, and the Arabic split of \mwiki~\citep{wiki40b}. For each dataset, we expand the model's vocabulary with all multi-token words that appear at least $m$ times in the test set.\footnote{We use $m=1,5,50$ for \wikitext, \pubmed~and Arabic \mwiki~respectively.} We then learn the refinement matrices $W_E$ and $W_U$ using 20M tokens from the train set.\footnote{We use a sequence length of 512 and train on 10,000 sequences, taking up to 30 minutes on a single GPU.} 
We evaluate models in a next-word prediction setup and compare our approach against two baselines: the original model with an unmodified vocabulary~(\textsl{original}); and an expanded model that follows our framework but uses the mean embeddings of the word's tokens in $E$ and $U$ to initialize its new representation, instead of the detokenized representations $r$, following \citeauthor{gee-etal-2022-fast}~(\citeyear{gee-etal-2022-fast}; \textsl{mean embedding}). 
For evaluation, perplexity is not a suitable metric, as differences in vocabulary between the approaches skew perplexity scores and prevent a fair comparison. Instead, we measure the model's token-level top-1 accuracy when it is given each token's previous context.

\begin{table}[tb]
    \centering
    \resizebox{0.99\textwidth}{!}{
    \begin{tabular}{lc|c|cc|c|cc|c|cc}
\toprule
 & & \multicolumn{3}{|c|}{\wikitext} & \multicolumn{3}{c|}{\pubmed} & \multicolumn{3}{c}{\mwiki-Arabic} \\
& & Original & Mean Embed. & Ours & Original & Mean Embed. & Ours & Original & Mean Embed. & Ours \\

\midrule
\multirow{2}{*}{New words} & \textsl{new} token & -- & 0.071 & 0.171 & -- & 0.123 & 0.180 & -- & 0.117 & 0.402 \\
 &  \textsl{original} or \textsl{new} token & 0.322 & 0.178 & 0.284 & 0.280 & 0.171 & 0.259 & 0.413 & 0.119 & 0.414  \\
\midrule
All words & &  0.522 &  0.473 &  0.519 &  0.517 &  0.479 &  0.511 &  0.535 &  0.211 &  0.532 \\
\bottomrule
\end{tabular}
    }
    \caption{Token-level accuracy of finetuning-free vocabulary expansion for Llama2-7B on three datasets. We compare our method using detokenized representations~(\emph{Ours}) to the original model~(\emph{Original}) and to an expansion baseline using the word's average token embeddings from $E$ and $U$~(\emph{Mean Embed.}; \citealt{gee-etal-2022-fast}). Accuracy is reported for newly added words, where we distinguish between correctly predicting the new tokens (top row) and predicting either the word's original first token, as the unexpanded model would, or its new token (middle row). We also report overall performance on all tokens (last row).
    Our method enables the frozen model to use the newly added input and output embeddings, while maintaining overall model performance.}
    \label{tab:vocab_expansion}
\end{table}

Our results~(\Cref{tab:vocab_expansion}) indicate that models can generalize to new vocabulary entries surprisingly well when these are initialized with their own detokenized representations. Unlike the \textsl{mean embeddings} expansion baseline, which both struggles with using new tokens and degrades overall performance, our method allows the model to successfully integrate new vocabulary words while preserving its accuracy on existing words. This effect is particularly pronounced in Arabic \mwiki, where the model almost always selects new tokens instead of the original tokenization (40.2\% vs 41.3\%), demonstrating the potential of our approach for multilingual and domain-specific applications. 

Our method also shows potential to reduce the number of tokens processed during encoding and inference: across the three datasets, we observe a reduction in the total number of tokens processed during encoding by 10.5\% to 14.5\% (see~\Cref{tab:token_savings} in~\cref{sec:vocab_expansion_efficency}), while maintaining model performance.
In future work, we will further explore the application of our framework to multilingual adaptation~\citep{petrov2024language, alabi-etal-2022-adapting} and continual domain-adaptive pretraining~\citep{ke2023continual, yildiz2024investigating, gururangan-etal-2020-dont}.

\section{Conclusion}
The ability of LLMs to comprehend and generate language relies on intricate internal processes, and understanding these mechanisms is crucial for improving model performance and efficiency. In this work, we unraveled the word detokenization process, shedding light on how models internally transform fragmented sub-word tokens into coherent word representations formed at the last token.
Our results indicate that this mechanism manifests in early to middle layers, where models attempt to reconstruct words by mapping them to an inner lexicon using their FFN layers. We provided evidence this lexicon is more exhaustive than the tokenizer's vocabulary, and could help models to recognize words even amidst noise.

Our work also unlocks practical avenues for optimizing tokenization, as well as the speed and cost of inference. We demonstrated one such application and presented a finetuning-free method to expand the vocabulary of LLMs. We hope our work paves the way for more efficient and versatile models.

\section*{Acknowledgments}
This work was supported in part by the Israel Science Foundation (grant no. 2045/21) and by NSF-BSF grant 2020793. We are grateful to Iddo Yosha for his valuable insights and thank the anonymous reviewers for their constructive feedback.

\bibliography{references}

\begin{thebibliography}{68}
\providecommand{\natexlab}[1]{#1}
\providecommand{\url}[1]{\texttt{#1}}
\expandafter\ifx\csname urlstyle\endcsname\relax
  \providecommand{\doi}[1]{doi: #1}\else
  \providecommand{\doi}{doi: \begingroup \urlstyle{rm}\Url}\fi

\bibitem[Ahia et~al.(2023)Ahia, Kumar, Gonen, Kasai, Mortensen, Smith, and Tsvetkov]{ahia-etal-2023-languages}
Orevaoghene Ahia, Sachin Kumar, Hila Gonen, Jungo Kasai, David Mortensen, Noah Smith, and Yulia Tsvetkov.
\newblock Do all languages cost the same? tokenization in the era of commercial language models.
\newblock In Houda Bouamor, Juan Pino, and Kalika Bali (eds.), \emph{Proceedings of the 2023 Conference on Empirical Methods in Natural Language Processing}, pp.\  9904--9923, Singapore, December 2023. Association for Computational Linguistics.
\newblock \doi{10.18653/v1/2023.emnlp-main.614}.
\newblock URL \url{https://aclanthology.org/2023.emnlp-main.614}.

\bibitem[AI et~al.(2024)AI, :, Young, Chen, Li, Huang, Zhang, Zhang, Li, Zhu, Chen, Chang, Yu, Liu, Liu, Yue, Yang, Yang, Yu, Xie, Huang, Hu, Ren, Niu, Nie, Xu, Liu, Wang, Cai, Gu, Liu, and Dai]{ai2024yiopenfoundationmodels}
01. AI, :, Alex Young, Bei Chen, Chao Li, Chengen Huang, Ge~Zhang, Guanwei Zhang, Heng Li, Jiangcheng Zhu, Jianqun Chen, Jing Chang, Kaidong Yu, Peng Liu, Qiang Liu, Shawn Yue, Senbin Yang, Shiming Yang, Tao Yu, Wen Xie, Wenhao Huang, Xiaohui Hu, Xiaoyi Ren, Xinyao Niu, Pengcheng Nie, Yuchi Xu, Yudong Liu, Yue Wang, Yuxuan Cai, Zhenyu Gu, Zhiyuan Liu, and Zonghong Dai.
\newblock Yi: Open foundation models by 01.ai, 2024.
\newblock URL \url{https://arxiv.org/abs/2403.04652}.

\bibitem[Aitchison(2012)]{aitchison2012words}
Jean Aitchison.
\newblock \emph{Words in the mind: An introduction to the mental lexicon}.
\newblock John Wiley \& Sons, 2012.

\bibitem[Alabi et~al.(2022)Alabi, Adelani, Mosbach, and Klakow]{alabi-etal-2022-adapting}
Jesujoba~O. Alabi, David~Ifeoluwa Adelani, Marius Mosbach, and Dietrich Klakow.
\newblock Adapting pre-trained language models to {A}frican languages via multilingual adaptive fine-tuning.
\newblock In Nicoletta Calzolari, Chu-Ren Huang, Hansaem Kim, James Pustejovsky, Leo Wanner, Key-Sun Choi, Pum-Mo Ryu, Hsin-Hsi Chen, Lucia Donatelli, Heng Ji, Sadao Kurohashi, Patrizia Paggio, Nianwen Xue, Seokhwan Kim, Younggyun Hahm, Zhong He, Tony~Kyungil Lee, Enrico Santus, Francis Bond, and Seung-Hoon Na (eds.), \emph{Proceedings of the 29th International Conference on Computational Linguistics}, pp.\  4336--4349, Gyeongju, Republic of Korea, October 2022. International Committee on Computational Linguistics.
\newblock URL \url{https://aclanthology.org/2022.coling-1.382/}.

\bibitem[Batsuren et~al.(2024)Batsuren, Vylomova, Dankers, Delgerbaatar, Uzan, Pinter, and Bella]{batsuren2024evaluating}
Khuyagbaatar Batsuren, Ekaterina Vylomova, Verna Dankers, Tsetsuukhei Delgerbaatar, Omri Uzan, Yuval Pinter, and G{\'a}bor Bella.
\newblock Evaluating subword tokenization: Alien subword composition and oov generalization challenge.
\newblock \emph{arXiv preprint arXiv:2404.13292}, 2024.

\bibitem[Bauwens \& Delobelle(2024)Bauwens and Delobelle]{bpe_knockout}
Thomas Bauwens and Pieter Delobelle.
\newblock Bpe-knockout: Pruning pre-existing bpe tokenisers with backwards-compatible morphological semi-supervision.
\newblock \emph{Conference of the North American Chapter of the Association for Computational Linguistics: Human Language Technologies}, 2024.

\bibitem[Belrose et~al.(2023)Belrose, Furman, Smith, Halawi, Ostrovsky, McKinney, Biderman, and Steinhardt]{belrose2023elicitinglatentpredictionstransformers}
Nora Belrose, Zach Furman, Logan Smith, Danny Halawi, Igor Ostrovsky, Lev McKinney, Stella Biderman, and Jacob Steinhardt.
\newblock Eliciting latent predictions from transformers with the tuned lens, 2023.
\newblock URL \url{https://arxiv.org/abs/2303.08112}.

\bibitem[Ben~Artzy \& Schwartz(2024)Ben~Artzy and Schwartz]{artzy-schwartz-2024-attend}
Amit Ben~Artzy and Roy Schwartz.
\newblock Attend first, consolidate later: On the importance of attention in different {LLM} layers.
\newblock In Yonatan Belinkov, Najoung Kim, Jaap Jumelet, Hosein Mohebbi, Aaron Mueller, and Hanjie Chen (eds.), \emph{Proceedings of the 7th BlackboxNLP Workshop: Analyzing and Interpreting Neural Networks for NLP}, pp.\  177--184, Miami, Florida, US, November 2024. Association for Computational Linguistics.
\newblock \doi{10.18653/v1/2024.blackboxnlp-1.10}.
\newblock URL \url{https://aclanthology.org/2024.blackboxnlp-1.10/}.

\bibitem[Bostrom \& Durrett(2020)Bostrom and Durrett]{bostrom-durrett-2020-byte}
Kaj Bostrom and Greg Durrett.
\newblock Byte pair encoding is suboptimal for language model pretraining.
\newblock In Trevor Cohn, Yulan He, and Yang Liu (eds.), \emph{Findings of the Association for Computational Linguistics: EMNLP 2020}, pp.\  4617--4624, Online, November 2020. Association for Computational Linguistics.
\newblock \doi{10.18653/v1/2020.findings-emnlp.414}.
\newblock URL \url{https://aclanthology.org/2020.findings-emnlp.414}.

\bibitem[Cao et~al.(2023)Cao, Kojima, Matsuo, and Iwasawa]{cao-etal-2023-unnatural}
Qi~Cao, Takeshi Kojima, Yutaka Matsuo, and Yusuke Iwasawa.
\newblock Unnatural error correction: {GPT}-4 can almost perfectly handle unnatural scrambled text.
\newblock In Houda Bouamor, Juan Pino, and Kalika Bali (eds.), \emph{Proceedings of the 2023 Conference on Empirical Methods in Natural Language Processing}, pp.\  8898--8913, Singapore, December 2023. Association for Computational Linguistics.
\newblock \doi{10.18653/v1/2023.emnlp-main.550}.
\newblock URL \url{https://aclanthology.org/2023.emnlp-main.550}.

\bibitem[Church(2020)]{church2020emerging}
Kenneth~Ward Church.
\newblock Emerging trends: Subwords, seriously?
\newblock \emph{Natural Language Engineering}, 26\penalty0 (3):\penalty0 375--382, 2020.

\bibitem[Conneau et~al.(2018)Conneau, Kruszewski, Lample, Barrault, and Baroni]{Probing}
Alexis Conneau, German Kruszewski, Guillaume Lample, Lo{\"\i}c Barrault, and Marco Baroni.
\newblock What you can cram into a single {\$}{\&}!{\#}* vector: Probing sentence embeddings for linguistic properties.
\newblock In Iryna Gurevych and Yusuke Miyao (eds.), \emph{Proceedings of the 56th Annual Meeting of the Association for Computational Linguistics (Volume 1: Long Papers)}, pp.\  2126--2136, Melbourne, Australia, July 2018. Association for Computational Linguistics.
\newblock \doi{10.18653/v1/P18-1198}.
\newblock URL \url{https://aclanthology.org/P18-1198}.

\bibitem[Correia et~al.(2019)Correia, Niculae, and Martins]{correia-etal-2019-adaptively}
Gon{\c{c}}alo~M. Correia, Vlad Niculae, and Andr{\'e} F.~T. Martins.
\newblock Adaptively sparse transformers.
\newblock In Kentaro Inui, Jing Jiang, Vincent Ng, and Xiaojun Wan (eds.), \emph{Proceedings of the 2019 Conference on Empirical Methods in Natural Language Processing and the 9th International Joint Conference on Natural Language Processing (EMNLP-IJCNLP)}, pp.\  2174--2184, Hong Kong, China, November 2019. Association for Computational Linguistics.
\newblock \doi{10.18653/v1/D19-1223}.
\newblock URL \url{https://aclanthology.org/D19-1223/}.

\bibitem[Dai et~al.(2022)Dai, Dong, Hao, Sui, Chang, and Wei]{dai-etal-2022-knowledge}
Damai Dai, Li~Dong, Yaru Hao, Zhifang Sui, Baobao Chang, and Furu Wei.
\newblock Knowledge neurons in pretrained transformers.
\newblock In Smaranda Muresan, Preslav Nakov, and Aline Villavicencio (eds.), \emph{Proceedings of the 60th Annual Meeting of the Association for Computational Linguistics (Volume 1: Long Papers)}, pp.\  8493--8502, Dublin, Ireland, May 2022. Association for Computational Linguistics.
\newblock \doi{10.18653/v1/2022.acl-long.581}.
\newblock URL \url{https://aclanthology.org/2022.acl-long.581}.

\bibitem[Dar et~al.(2023)Dar, Geva, Gupta, and Berant]{dar-etal-2023-analyzing}
Guy Dar, Mor Geva, Ankit Gupta, and Jonathan Berant.
\newblock Analyzing transformers in embedding space.
\newblock In Anna Rogers, Jordan Boyd-Graber, and Naoaki Okazaki (eds.), \emph{Proceedings of the 61st Annual Meeting of the Association for Computational Linguistics (Volume 1: Long Papers)}, pp.\  16124--16170, Toronto, Canada, July 2023. Association for Computational Linguistics.
\newblock \doi{10.18653/v1/2023.acl-long.893}.
\newblock URL \url{https://aclanthology.org/2023.acl-long.893}.

\bibitem[Dubey et~al.(2024)Dubey, Jauhri, Pandey, Kadian, Al-Dahle, Letman, Mathur, Schelten, Yang, Fan, et~al.]{dubey2024llama3herdmodels}
Abhimanyu Dubey, Abhinav Jauhri, Abhinav Pandey, Abhishek Kadian, Ahmad Al-Dahle, Aiesha Letman, Akhil Mathur, Alan Schelten, Amy Yang, Angela Fan, et~al.
\newblock The llama 3 herd of models.
\newblock \emph{arXiv preprint arXiv:2407.21783}, 2024.

\bibitem[Durrani et~al.(2020)Durrani, Sajjad, Dalvi, and Belinkov]{durrani-etal-2020-analyzing}
Nadir Durrani, Hassan Sajjad, Fahim Dalvi, and Yonatan Belinkov.
\newblock Analyzing individual neurons in pre-trained language models.
\newblock In Bonnie Webber, Trevor Cohn, Yulan He, and Yang Liu (eds.), \emph{Proceedings of the 2020 Conference on Empirical Methods in Natural Language Processing (EMNLP)}, pp.\  4865--4880, Online, November 2020. Association for Computational Linguistics.
\newblock \doi{10.18653/v1/2020.emnlp-main.395}.
\newblock URL \url{https://aclanthology.org/2020.emnlp-main.395}.

\bibitem[Elhage et~al.(2022)Elhage, Hume, Olsson, Nanda, Henighan, Johnston, ElShowk, Joseph, DasSarma, Mann, Hernandez, Askell, Ndousse, Jones, Drain, Chen, Bai, Ganguli, Lovitt, Hatfield-Dodds, Kernion, Conerly, Kravec, Fort, Kadavath, Jacobson, Tran-Johnson, Kaplan, Clark, Brown, McCandlish, Amodei, and Olah]{elhage2022solu}
Nelson Elhage, Tristan Hume, Catherine Olsson, Neel Nanda, Tom Henighan, Scott Johnston, Sheer ElShowk, Nicholas Joseph, Nova DasSarma, Ben Mann, Danny Hernandez, Amanda Askell, Kamal Ndousse, Andy Jones, Dawn Drain, Anna Chen, Yuntao Bai, Deep Ganguli, Liane Lovitt, Zac Hatfield-Dodds, Jackson Kernion, Tom Conerly, Shauna Kravec, Stanislav Fort, Saurav Kadavath, Josh Jacobson, Eli Tran-Johnson, Jared Kaplan, Jack Clark, Tom Brown, Sam McCandlish, Dario Amodei, and Christopher Olah.
\newblock Softmax linear units.
\newblock \emph{Transformer Circuits Thread}, 2022.
\newblock https://transformer-circuits.pub/2022/solu/index.html.

\bibitem[Ferrando \& Voita(2024)Ferrando and Voita]{ferrando2024information}
Javier Ferrando and Elena Voita.
\newblock Information flow routes: Automatically interpreting language models at scale.
\newblock \emph{arXiv preprint arXiv:2403.00824}, 2024.

\bibitem[Feucht et~al.(2024)Feucht, Atkinson, Wallace, and Bau]{feucht2024tokenerasurefootprintimplicit}
Sheridan Feucht, David Atkinson, Byron~C Wallace, and David Bau.
\newblock Token erasure as a footprint of implicit vocabulary items in {LLM}s.
\newblock In Yaser Al-Onaizan, Mohit Bansal, and Yun-Nung Chen (eds.), \emph{Proceedings of the 2024 Conference on Empirical Methods in Natural Language Processing}, pp.\  9727--9739, Miami, Florida, USA, November 2024. Association for Computational Linguistics.
\newblock \doi{10.18653/v1/2024.emnlp-main.543}.
\newblock URL \url{https://aclanthology.org/2024.emnlp-main.543/}.

\bibitem[Frisch et~al.(2000)Frisch, Large, and Pisoni]{Frisch2000PerceptionOW}
Stefan~A. Frisch, Nathan~R. Large, and David~B. Pisoni.
\newblock Perception of wordlikeness: Effects of segment probability and length on the processing of nonwords.
\newblock \emph{Journal of memory and language}, 42 4:\penalty0 481--496, 2000.
\newblock URL \url{https://api.semanticscholar.org/CorpusID:1736965}.

\bibitem[Gee et~al.(2022)Gee, Zugarini, Rigutini, and Torroni]{gee-etal-2022-fast}
Leonidas Gee, Andrea Zugarini, Leonardo Rigutini, and Paolo Torroni.
\newblock Fast vocabulary transfer for language model compression.
\newblock In Yunyao Li and Angeliki Lazaridou (eds.), \emph{Proceedings of the 2022 Conference on Empirical Methods in Natural Language Processing: Industry Track}, pp.\  409--416, Abu Dhabi, UAE, December 2022. Association for Computational Linguistics.
\newblock \doi{10.18653/v1/2022.emnlp-industry.41}.
\newblock URL \url{https://aclanthology.org/2022.emnlp-industry.41/}.

\bibitem[Gerlach \& Font-Clos(2018)Gerlach and Font-Clos]{gerlach2018standardizedprojectgutenbergcorpus}
Martin Gerlach and Francesc Font-Clos.
\newblock A standardized project gutenberg corpus for statistical analysis of natural language and quantitative linguistics, 2018.
\newblock URL \url{https://arxiv.org/abs/1812.08092}.

\bibitem[Geva et~al.(2021)Geva, Schuster, Berant, and Levy]{geva-etal-2021-transformer}
Mor Geva, Roei Schuster, Jonathan Berant, and Omer Levy.
\newblock Transformer feed-forward layers are key-value memories.
\newblock In Marie-Francine Moens, Xuanjing Huang, Lucia Specia, and Scott Wen-tau Yih (eds.), \emph{Proceedings of the 2021 Conference on Empirical Methods in Natural Language Processing}, pp.\  5484--5495, Online and Punta Cana, Dominican Republic, November 2021. Association for Computational Linguistics.
\newblock \doi{10.18653/v1/2021.emnlp-main.446}.
\newblock URL \url{https://aclanthology.org/2021.emnlp-main.446}.

\bibitem[Geva et~al.(2022)Geva, Caciularu, Wang, and Goldberg]{geva-etal-2022-transformer}
Mor Geva, Avi Caciularu, Kevin Wang, and Yoav Goldberg.
\newblock Transformer feed-forward layers build predictions by promoting concepts in the vocabulary space.
\newblock In Yoav Goldberg, Zornitsa Kozareva, and Yue Zhang (eds.), \emph{Proceedings of the 2022 Conference on Empirical Methods in Natural Language Processing}, pp.\  30--45, Abu Dhabi, United Arab Emirates, December 2022. Association for Computational Linguistics.
\newblock \doi{10.18653/v1/2022.emnlp-main.3}.
\newblock URL \url{https://aclanthology.org/2022.emnlp-main.3}.

\bibitem[Geva et~al.(2023)Geva, Bastings, Filippova, and Globerson]{geva-etal-2023-dissecting}
Mor Geva, Jasmijn Bastings, Katja Filippova, and Amir Globerson.
\newblock Dissecting recall of factual associations in auto-regressive language models.
\newblock In Houda Bouamor, Juan Pino, and Kalika Bali (eds.), \emph{Proceedings of the 2023 Conference on Empirical Methods in Natural Language Processing}, pp.\  12216--12235, Singapore, December 2023. Association for Computational Linguistics.
\newblock \doi{10.18653/v1/2023.emnlp-main.751}.
\newblock URL \url{https://aclanthology.org/2023.emnlp-main.751}.

\bibitem[Ghandeharioun et~al.(2024)Ghandeharioun, Caciularu, Pearce, Dixon, and Geva]{Ghandeharioun2024PatchscopesAU}
Asma Ghandeharioun, Avi Caciularu, Adam Pearce, Lucas Dixon, and Mor Geva.
\newblock Patchscopes: A unifying framework for inspecting hidden representations of language models.
\newblock In \emph{Proc. of ICML}, volume abs/2401.06102, 2024.
\newblock URL \url{https://arxiv.org/abs/2401.06102}.

\bibitem[Guo et~al.(2020)Guo, Dai, Vrande{\v{c}}i{\'c}, and Al-Rfou]{wiki40b}
Mandy Guo, Zihang Dai, Denny Vrande{\v{c}}i{\'c}, and Rami Al-Rfou.
\newblock {W}iki-40{B}: Multilingual language model dataset.
\newblock In Nicoletta Calzolari, Fr{\'e}d{\'e}ric B{\'e}chet, Philippe Blache, Khalid Choukri, Christopher Cieri, Thierry Declerck, Sara Goggi, Hitoshi Isahara, Bente Maegaard, Joseph Mariani, H{\'e}l{\`e}ne Mazo, Asuncion Moreno, Jan Odijk, and Stelios Piperidis (eds.), \emph{Proceedings of the Twelfth Language Resources and Evaluation Conference}, pp.\  2440--2452, Marseille, France, May 2020. European Language Resources Association.
\newblock ISBN 979-10-95546-34-4.
\newblock URL \url{https://aclanthology.org/2020.lrec-1.297/}.

\bibitem[Gurnee et~al.(2023)Gurnee, Nanda, Pauly, Harvey, Troitskii, and Bertsimas]{gurnee2023finding}
Wes Gurnee, Neel Nanda, Matthew Pauly, Katherine Harvey, Dmitrii Troitskii, and Dimitris Bertsimas.
\newblock Finding neurons in a haystack: Case studies with sparse probing.
\newblock \emph{Transactions on Machine Learning Research}, 2023.
\newblock ISSN 2835-8856.
\newblock URL \url{https://openreview.net/forum?id=JYs1R9IMJr}.

\bibitem[Gururangan et~al.(2020)Gururangan, Marasovi{\'c}, Swayamdipta, Lo, Beltagy, Downey, and Smith]{gururangan-etal-2020-dont}
Suchin Gururangan, Ana Marasovi{\'c}, Swabha Swayamdipta, Kyle Lo, Iz~Beltagy, Doug Downey, and Noah~A. Smith.
\newblock Don`t stop pretraining: Adapt language models to domains and tasks.
\newblock In Dan Jurafsky, Joyce Chai, Natalie Schluter, and Joel Tetreault (eds.), \emph{Proceedings of the 58th Annual Meeting of the Association for Computational Linguistics}, pp.\  8342--8360, Online, July 2020. Association for Computational Linguistics.
\newblock \doi{10.18653/v1/2020.acl-main.740}.
\newblock URL \url{https://aclanthology.org/2020.acl-main.740/}.

\bibitem[Hofmann et~al.(2022)Hofmann, Schuetze, and Pierrehumbert]{hofmann-etal-2022-embarrassingly}
Valentin Hofmann, Hinrich Schuetze, and Janet Pierrehumbert.
\newblock An embarrassingly simple method to mitigate undesirable properties of pretrained language model tokenizers.
\newblock In Smaranda Muresan, Preslav Nakov, and Aline Villavicencio (eds.), \emph{Proceedings of the 60th Annual Meeting of the Association for Computational Linguistics (Volume 2: Short Papers)}, pp.\  385--393, Dublin, Ireland, May 2022. Association for Computational Linguistics.
\newblock \doi{10.18653/v1/2022.acl-short.43}.
\newblock URL \url{https://aclanthology.org/2022.acl-short.43}.

\bibitem[Jiang et~al.(2023)Jiang, Sablayrolles, Mensch, Bamford, Chaplot, de~las Casas, Bressand, Lengyel, Lample, Saulnier, Lavaud, Lachaux, Stock, Scao, Lavril, Wang, Lacroix, and Sayed]{jiang2023mistral7b}
Albert~Q. Jiang, Alexandre Sablayrolles, Arthur Mensch, Chris Bamford, Devendra~Singh Chaplot, Diego de~las Casas, Florian Bressand, Gianna Lengyel, Guillaume Lample, Lucile Saulnier, Lélio~Renard Lavaud, Marie-Anne Lachaux, Pierre Stock, Teven~Le Scao, Thibaut Lavril, Thomas Wang, Timothée Lacroix, and William~El Sayed.
\newblock Mistral 7b, 2023.
\newblock URL \url{https://arxiv.org/abs/2310.06825}.

\bibitem[Kaushal \& Mahowald(2022)Kaushal and Mahowald]{kaushal-mahowald-2022-tokens}
Ayush Kaushal and Kyle Mahowald.
\newblock What do tokens know about their characters and how do they know it?
\newblock In Marine Carpuat, Marie-Catherine de~Marneffe, and Ivan~Vladimir Meza~Ruiz (eds.), \emph{Proceedings of the 2022 Conference of the North American Chapter of the Association for Computational Linguistics: Human Language Technologies}, pp.\  2487--2507, Seattle, United States, July 2022. Association for Computational Linguistics.
\newblock \doi{10.18653/v1/2022.naacl-main.179}.
\newblock URL \url{https://aclanthology.org/2022.naacl-main.179}.

\bibitem[Ke et~al.(2023)Ke, Shao, Lin, Konishi, Kim, and Liu]{ke2023continual}
Zixuan Ke, Yijia Shao, Haowei Lin, Tatsuya Konishi, Gyuhak Kim, and Bing Liu.
\newblock Continual pre-training of language models.
\newblock In \emph{The Eleventh International Conference on Learning Representations}, 2023.
\newblock URL \url{https://openreview.net/forum?id=m_GDIItaI3o}.

\bibitem[Kim et~al.(2024)Kim, Choi, and Jeong]{kim2024efficient}
Seungduk Kim, Seungtaek Choi, and Myeongho Jeong.
\newblock Efficient and effective vocabulary expansion towards multilingual large language models.
\newblock \emph{arXiv preprint arXiv:2402.14714}, 2024.

\bibitem[Klein \& Tsarfaty(2020)Klein and Tsarfaty]{klein2020getting}
Stav Klein and Reut Tsarfaty.
\newblock Getting the\#\# life out of living: How adequate are word-pieces for modelling complex morphology?
\newblock In \emph{Proceedings of the 17th SIGMORPHON workshop on computational research in phonetics, phonology, and morphology}, pp.\  204--209, 2020.

\bibitem[Kudo(2018)]{kudo-2018-subword}
Taku Kudo.
\newblock Subword regularization: Improving neural network translation models with multiple subword candidates.
\newblock In Iryna Gurevych and Yusuke Miyao (eds.), \emph{Proceedings of the 56th Annual Meeting of the Association for Computational Linguistics (Volume 1: Long Papers)}, pp.\  66--75, Melbourne, Australia, July 2018. Association for Computational Linguistics.
\newblock \doi{10.18653/v1/P18-1007}.
\newblock URL \url{https://aclanthology.org/P18-1007}.

\bibitem[Lad et~al.(2024)Lad, Gurnee, and Tegmark]{lad2024remarkablerobustnessllmsstages}
Vedang Lad, Wes Gurnee, and Max Tegmark.
\newblock The remarkable robustness of llms: Stages of inference?, 2024.
\newblock URL \url{https://arxiv.org/abs/2406.19384}.

\bibitem[Marslen-Wilson et~al.(1994)Marslen-Wilson, Tyler, Waksler, and Older]{marslen1994morphology}
William Marslen-Wilson, Lorraine~K Tyler, Rachelle Waksler, and Lianne Older.
\newblock Morphology and meaning in the english mental lexicon.
\newblock \emph{Psychological review}, 101\penalty0 (1):\penalty0 3, 1994.

\bibitem[Meng et~al.(2022{\natexlab{a}})Meng, Richer, Tehrani, La, Kim, Ayers, and Heidar-Zadeh]{meng2022procrustes}
Fanwang Meng, Michael Richer, Alireza Tehrani, Jonathan La, Taewon~David Kim, Paul~W Ayers, and Farnaz Heidar-Zadeh.
\newblock Procrustes: A python library to find transformations that maximize the similarity between matrices.
\newblock \emph{Computer Physics Communications}, 276:\penalty0 108334, 2022{\natexlab{a}}.

\bibitem[Meng et~al.(2022{\natexlab{b}})Meng, Bau, Andonian, and Belinkov]{meng2022locating}
Kevin Meng, David Bau, Alex Andonian, and Yonatan Belinkov.
\newblock Locating and editing factual associations in gpt.
\newblock \emph{Advances in Neural Information Processing Systems}, 35:\penalty0 17359--17372, 2022{\natexlab{b}}.

\bibitem[Merity et~al.(2017)Merity, Xiong, Bradbury, and Socher]{wikitext}
Stephen Merity, Caiming Xiong, James Bradbury, and Richard Socher.
\newblock Pointer sentinel mixture models.
\newblock In \emph{International Conference on Learning Representations}, 2017.
\newblock URL \url{https://openreview.net/forum?id=Byj72udxe}.

\bibitem[Merullo et~al.(2024)Merullo, Eickhoff, and Pavlick]{merullo-etal-2024-language}
Jack Merullo, Carsten Eickhoff, and Ellie Pavlick.
\newblock Language models implement simple {W}ord2{V}ec-style vector arithmetic.
\newblock In Kevin Duh, Helena Gomez, and Steven Bethard (eds.), \emph{Proceedings of the 2024 Conference of the North American Chapter of the Association for Computational Linguistics: Human Language Technologies (Volume 1: Long Papers)}, pp.\  5030--5047, Mexico City, Mexico, June 2024. Association for Computational Linguistics.
\newblock \doi{10.18653/v1/2024.naacl-long.281}.
\newblock URL \url{https://aclanthology.org/2024.naacl-long.281}.

\bibitem[nostalgebraist(2020)]{nostalgebraist2020logitlens}
nostalgebraist.
\newblock interpreting {GPT}: the logit lens.
\newblock \emph{LessWrong}, 2020.
\newblock URL \url{https://www.lesswrong.com/posts/AcKRB8wDpdaN6v6ru/interpreting-gpt-the-logit-lens}.

\bibitem[Pal et~al.(2023)Pal, Sun, Yuan, Wallace, and Bau]{pal-etal-2023-future}
Koyena Pal, Jiuding Sun, Andrew Yuan, Byron Wallace, and David Bau.
\newblock Future lens: Anticipating subsequent tokens from a single hidden state.
\newblock In Jing Jiang, David Reitter, and Shumin Deng (eds.), \emph{Proceedings of the 27th Conference on Computational Natural Language Learning (CoNLL)}, pp.\  548--560, Singapore, December 2023. Association for Computational Linguistics.
\newblock \doi{10.18653/v1/2023.conll-1.37}.
\newblock URL \url{https://aclanthology.org/2023.conll-1.37}.

\bibitem[Park et~al.(2024)Park, Choe, Jiang, and Veitch]{park2024geometry}
Kiho Park, Yo~Joong Choe, Yibo Jiang, and Victor Veitch.
\newblock The geometry of categorical and hierarchical concepts in large language models.
\newblock In \emph{ICML 2024 Workshop on Mechanistic Interpretability}, 2024.
\newblock URL \url{https://openreview.net/forum?id=KXuYjuBzKo}.

\bibitem[Park et~al.(2025)Park, Choe, Jiang, and Veitch]{LRH}
Kiho Park, Yo~Joong Choe, Yibo Jiang, and Victor Veitch.
\newblock The representation geometry of features and hierarchy in large language models.
\newblock In \emph{The Thirteenth International Conference on Learning Representations}, 2025.
\newblock URL \url{https://openreview.net/forum?id=bVTM2QKYuA}.

\bibitem[Petrov et~al.(2023)Petrov, La~Malfa, Torr, and Bibi]{petrov2024language}
Aleksandar Petrov, Emanuele La~Malfa, Philip Torr, and Adel Bibi.
\newblock Language model tokenizers introduce unfairness between languages.
\newblock \emph{Advances in Neural Information Processing Systems}, 36, 2023.

\bibitem[Provilkov et~al.(2020)Provilkov, Emelianenko, and Voita]{provilkov-etal-2020-bpe}
Ivan Provilkov, Dmitrii Emelianenko, and Elena Voita.
\newblock {BPE}-dropout: Simple and effective subword regularization.
\newblock In Dan Jurafsky, Joyce Chai, Natalie Schluter, and Joel Tetreault (eds.), \emph{Proceedings of the 58th Annual Meeting of the Association for Computational Linguistics}, pp.\  1882--1892, Online, July 2020. Association for Computational Linguistics.
\newblock \doi{10.18653/v1/2020.acl-main.170}.
\newblock URL \url{https://aclanthology.org/2020.acl-main.170}.

\bibitem[Rastle et~al.(2002)Rastle, Harrington, and Coltheart]{rastle2002arc}
Kathleen Rastle, Jonathan Harrington, and Max Coltheart.
\newblock 358,534 nonwords: the arc nonword database.
\newblock \emph{The Quarterly journal of experimental psychology. A, Human experimental psychology}, 55:\penalty0 1339--62, 11 2002.
\newblock \doi{10.1080/02724980244000099}.

\bibitem[Sajjad et~al.(2022)Sajjad, Durrani, Dalvi, Alam, Khan, and Xu]{sajjad-etal-2022-analyzing}
Hassan Sajjad, Nadir Durrani, Fahim Dalvi, Firoj Alam, Abdul Khan, and Jia Xu.
\newblock Analyzing encoded concepts in transformer language models.
\newblock In Marine Carpuat, Marie-Catherine de~Marneffe, and Ivan~Vladimir Meza~Ruiz (eds.), \emph{Proceedings of the 2022 Conference of the North American Chapter of the Association for Computational Linguistics: Human Language Technologies}, pp.\  3082--3101, Seattle, United States, July 2022. Association for Computational Linguistics.
\newblock \doi{10.18653/v1/2022.naacl-main.225}.
\newblock URL \url{https://aclanthology.org/2022.naacl-main.225/}.

\bibitem[Schmidt et~al.(2024)Schmidt, Reddy, Zhang, Alameddine, Uzan, Pinter, and Tanner]{schmidt2024tokenization}
Craig~W Schmidt, Varshini Reddy, Haoran Zhang, Alec Alameddine, Omri Uzan, Yuval Pinter, and Chris Tanner.
\newblock Tokenization is more than compression.
\newblock \emph{arXiv preprint arXiv:2402.18376}, 2024.

\bibitem[Sch{\"o}nemann(1966)]{schonemann1966generalized}
Peter~H Sch{\"o}nemann.
\newblock A generalized solution of the orthogonal procrustes problem.
\newblock \emph{Psychometrika}, 31\penalty0 (1):\penalty0 1--10, 1966.

\bibitem[Sengupta et~al.(2023)Sengupta, Sahu, Jia, Katipomu, Li, Koto, Marshall, Gosal, Liu, Chen, Afzal, Kamboj, Pandit, Pal, Pradhan, Mujahid, Baali, Han, Bsharat, Aji, Shen, Liu, Vassilieva, Hestness, Hock, Feldman, Lee, Jackson, Ren, Nakov, Baldwin, and Xing]{sengupta2023jais}
Neha Sengupta, Sunil~Kumar Sahu, Bokang Jia, Satheesh Katipomu, Haonan Li, Fajri Koto, William Marshall, Gurpreet Gosal, Cynthia Liu, Zhiming Chen, Osama~Mohammed Afzal, Samta Kamboj, Onkar Pandit, Rahul Pal, Lalit Pradhan, Zain~Muhammad Mujahid, Massa Baali, Xudong Han, Sondos~Mahmoud Bsharat, Alham~Fikri Aji, Zhiqiang Shen, Zhengzhong Liu, Natalia Vassilieva, Joel Hestness, Andy Hock, Andrew Feldman, Jonathan Lee, Andrew Jackson, Hector~Xuguang Ren, Preslav Nakov, Timothy Baldwin, and Eric Xing.
\newblock Jais and jais-chat: Arabic-centric foundation and instruction-tuned open generative large language models, 2023.
\newblock URL \url{https://arxiv.org/abs/2308.16149}.
\newblock {arXiv}:2308.16149.

\bibitem[Sennrich et~al.(2016)Sennrich, Haddow, and Birch]{sennrich-etal-2016-neural}
Rico Sennrich, Barry Haddow, and Alexandra Birch.
\newblock Neural machine translation of rare words with subword units.
\newblock In Katrin Erk and Noah~A. Smith (eds.), \emph{Proceedings of the 54th Annual Meeting of the Association for Computational Linguistics (Volume 1: Long Papers)}, pp.\  1715--1725, Berlin, Germany, August 2016. Association for Computational Linguistics.
\newblock \doi{10.18653/v1/P16-1162}.
\newblock URL \url{https://aclanthology.org/P16-1162}.

\bibitem[Tenney et~al.(2019)Tenney, Das, and Pavlick]{BERT_Rediscovers}
Ian Tenney, Dipanjan Das, and Ellie Pavlick.
\newblock {BERT} rediscovers the classical {NLP} pipeline.
\newblock In Anna Korhonen, David Traum, and Llu{\'\i}s M{\`a}rquez (eds.), \emph{Proceedings of the 57th Annual Meeting of the Association for Computational Linguistics}, pp.\  4593--4601, Florence, Italy, July 2019. Association for Computational Linguistics.
\newblock \doi{10.18653/v1/P19-1452}.
\newblock URL \url{https://aclanthology.org/P19-1452}.

\bibitem[Todd et~al.(2024)Todd, Li, Sharma, Mueller, Wallace, and Bau]{todd2024function}
Eric Todd, Millicent Li, Arnab~Sen Sharma, Aaron Mueller, Byron~C Wallace, and David Bau.
\newblock Function vectors in large language models.
\newblock In \emph{The Twelfth International Conference on Learning Representations}, 2024.
\newblock URL \url{https://openreview.net/forum?id=AwyxtyMwaG}.

\bibitem[Touvron et~al.(2023)Touvron, Martin, Stone, Albert, Almahairi, Babaei, Bashlykov, Batra, Bhargava, Bhosale, Bikel, Blecher, Ferrer, Chen, Cucurull, Esiobu, Fernandes, Fu, Fu, Fuller, Gao, Goswami, Goyal, Hartshorn, Hosseini, Hou, Inan, Kardas, Kerkez, Khabsa, Kloumann, Korenev, Koura, Lachaux, Lavril, Lee, Liskovich, Lu, Mao, Martinet, Mihaylov, Mishra, Molybog, Nie, Poulton, Reizenstein, Rungta, Saladi, Schelten, Silva, Smith, Subramanian, Tan, Tang, Taylor, Williams, Kuan, Xu, Yan, Zarov, Zhang, Fan, Kambadur, Narang, Rodriguez, Stojnic, Edunov, and Scialom]{touvron2023llama2openfoundation}
Hugo Touvron, Louis Martin, Kevin Stone, Peter Albert, Amjad Almahairi, Yasmine Babaei, Nikolay Bashlykov, Soumya Batra, Prajjwal Bhargava, Shruti Bhosale, Dan Bikel, Lukas Blecher, Cristian~Canton Ferrer, Moya Chen, Guillem Cucurull, David Esiobu, Jude Fernandes, Jeremy Fu, Wenyin Fu, Brian Fuller, Cynthia Gao, Vedanuj Goswami, Naman Goyal, Anthony Hartshorn, Saghar Hosseini, Rui Hou, Hakan Inan, Marcin Kardas, Viktor Kerkez, Madian Khabsa, Isabel Kloumann, Artem Korenev, Punit~Singh Koura, Marie-Anne Lachaux, Thibaut Lavril, Jenya Lee, Diana Liskovich, Yinghai Lu, Yuning Mao, Xavier Martinet, Todor Mihaylov, Pushkar Mishra, Igor Molybog, Yixin Nie, Andrew Poulton, Jeremy Reizenstein, Rashi Rungta, Kalyan Saladi, Alan Schelten, Ruan Silva, Eric~Michael Smith, Ranjan Subramanian, Xiaoqing~Ellen Tan, Binh Tang, Ross Taylor, Adina Williams, Jian~Xiang Kuan, Puxin Xu, Zheng Yan, Iliyan Zarov, Yuchen Zhang, Angela Fan, Melanie Kambadur, Sharan Narang, Aurelien Rodriguez, Robert Stojnic, Sergey Edunov, and Thomas
  Scialom.
\newblock Llama 2: Open foundation and fine-tuned chat models, 2023.
\newblock URL \url{https://arxiv.org/abs/2307.09288}.

\bibitem[Vuli{\'c} et~al.(2020)Vuli{\'c}, Ponti, Litschko, Glava{\v{s}}, and Korhonen]{vulic-etal-2020-probing}
Ivan Vuli{\'c}, Edoardo~Maria Ponti, Robert Litschko, Goran Glava{\v{s}}, and Anna Korhonen.
\newblock Probing pretrained language models for lexical semantics.
\newblock In Bonnie Webber, Trevor Cohn, Yulan He, and Yang Liu (eds.), \emph{Proceedings of the 2020 Conference on Empirical Methods in Natural Language Processing (EMNLP)}, pp.\  7222--7240, Online, November 2020. Association for Computational Linguistics.
\newblock \doi{10.18653/v1/2020.emnlp-main.586}.
\newblock URL \url{https://aclanthology.org/2020.emnlp-main.586}.

\bibitem[Wu(2016)]{wu2016wordpiece}
Yonghui Wu.
\newblock Google’s neural machine translation system: Bridging the gap between human and machine translation.
\newblock \emph{arXiv preprint arXiv:1609.08144}, 2016.

\bibitem[Xiong et~al.(2024)Xiong, Jin, Lu, and Zhang]{pubmed}
Guangzhi Xiong, Qiao Jin, Zhiyong Lu, and Aidong Zhang.
\newblock Benchmarking retrieval-augmented generation for medicine.
\newblock In Lun-Wei Ku, Andre Martins, and Vivek Srikumar (eds.), \emph{Findings of the Association for Computational Linguistics: ACL 2024}, pp.\  6233--6251, Bangkok, Thailand, August 2024. Association for Computational Linguistics.
\newblock \doi{10.18653/v1/2024.findings-acl.372}.
\newblock URL \url{https://aclanthology.org/2024.findings-acl.372/}.

\bibitem[Yehezkel \& Pinter(2023)Yehezkel and Pinter]{yehezkel-pinter-2023-incorporating}
Shaked Yehezkel and Yuval Pinter.
\newblock Incorporating context into subword vocabularies.
\newblock In Andreas Vlachos and Isabelle Augenstein (eds.), \emph{Proceedings of the 17th Conference of the European Chapter of the Association for Computational Linguistics}, pp.\  623--635, Dubrovnik, Croatia, May 2023. Association for Computational Linguistics.
\newblock \doi{10.18653/v1/2023.eacl-main.45}.
\newblock URL \url{https://aclanthology.org/2023.eacl-main.45}.

\bibitem[Y{\i}ld{\i}z et~al.(2024)Y{\i}ld{\i}z, Ravichandran, Punia, Bethge, and Ermis]{yildiz2024investigating}
{\c{C}}a{\u{g}}atay Y{\i}ld{\i}z, Nishaanth~Kanna Ravichandran, Prishruit Punia, Matthias Bethge, and Beyza Ermis.
\newblock Investigating continual pretraining in large language models: Insights and implications.
\newblock \emph{arXiv preprint arXiv:2402.17400}, 2024.

\bibitem[Yom~Din et~al.(2024)Yom~Din, Karidi, Choshen, and Geva]{yom-din-etal-2024-jump}
Alexander Yom~Din, Taelin Karidi, Leshem Choshen, and Mor Geva.
\newblock Jump to conclusions: Short-cutting transformers with linear transformations.
\newblock In Nicoletta Calzolari, Min-Yen Kan, Veronique Hoste, Alessandro Lenci, Sakriani Sakti, and Nianwen Xue (eds.), \emph{Proceedings of the 2024 Joint International Conference on Computational Linguistics, Language Resources and Evaluation (LREC-COLING 2024)}, pp.\  9615--9625, Torino, Italia, May 2024. ELRA and ICCL.
\newblock URL \url{https://aclanthology.org/2024.lrec-main.840}.

\bibitem[Zhang \& Sennrich(2019)Zhang and Sennrich]{RMS}
Biao Zhang and Rico Sennrich.
\newblock Root mean square layer normalization.
\newblock \emph{Advances in Neural Information Processing Systems}, 32, 2019.

\bibitem[Zhang \& Nanda(2024)Zhang and Nanda]{zhang2024towards}
Fred Zhang and Neel Nanda.
\newblock Towards best practices of activation patching in language models: Metrics and methods.
\newblock In \emph{The Twelfth International Conference on Learning Representations}, 2024.
\newblock URL \url{https://openreview.net/forum?id=Hf17y6u9BC}.

\bibitem[Zhao et~al.(2024)Zhao, Zhang, Zhang, Gui, and Huang]{zhao2024llama}
Jun Zhao, Zhihao Zhang, Qi~Zhang, Tao Gui, and Xuanjing Huang.
\newblock Llama beyond english: An empirical study on language capability transfer.
\newblock \emph{arXiv preprint arXiv:2401.01055}, 2024.

\bibitem[Zouhar et~al.(2023)Zouhar, Meister, Gastaldi, Du, Sachan, and Cotterell]{zouhar-etal-2023-tokenization}
Vil{\'e}m Zouhar, Clara Meister, Juan Gastaldi, Li~Du, Mrinmaya Sachan, and Ryan Cotterell.
\newblock Tokenization and the noiseless channel.
\newblock In Anna Rogers, Jordan Boyd-Graber, and Naoaki Okazaki (eds.), \emph{Proceedings of the 61st Annual Meeting of the Association for Computational Linguistics (Volume 1: Long Papers)}, pp.\  5184--5207, Toronto, Canada, July 2023. Association for Computational Linguistics.
\newblock \doi{10.18653/v1/2023.acl-long.284}.
\newblock URL \url{https://aclanthology.org/2023.acl-long.284}.

\end{thebibliography}
\bibliographystyle{iclr2025_conference}

\newpage

\appendix
\section{Words vs.~Nonwords Analysis}
\label{sec:appendix_words_vs_nonwords}
To further analyze our results from~\cref{sec:word_vs_gibberish}, we conduct a failure analysis, 
focusing on false negatives (FN) and false positives (FP). 
Our analysis indicates that the former---valid words misclassified as gibberish---are 
often rare and complex words, suggesting that the model's internal 
vocabulary may lack representations for infrequent words. 
On the other hand, false positives---where gibberish is misclassified as a 
word---typically involves nonwords that closely resemble valid words, 
likely due to shared sub-word structures.
See~\cref{table:failures} for a few examples.

\begin{table}[!ht]
\centering
\begin{tabular}{@{}ccccc@{}}
\toprule
\textbf{Original Word}  & \textbf{Predicted Value} & \textbf{True Value} & \textbf{Status} \\ \midrule
\_Unitarianism & gibberish     & word     & FN  \\ 
\_killy        & gibberish     & word     & FN  \\ 
\_quadruropic  & word          & gibberish & FP  \\ 
\_nonwith      & word          & gibberish & FP  \\ \bottomrule
\end{tabular}
\caption{Examples of false negative (FN) and false positive (FP) for the word vs.~nonword experiment~(\cref{sec:word_vs_gibberish}).}
\label{table:failures}
\end{table}

\begin{figure}[!ht]
    \centering
    \includegraphics[width=0.8\textwidth]{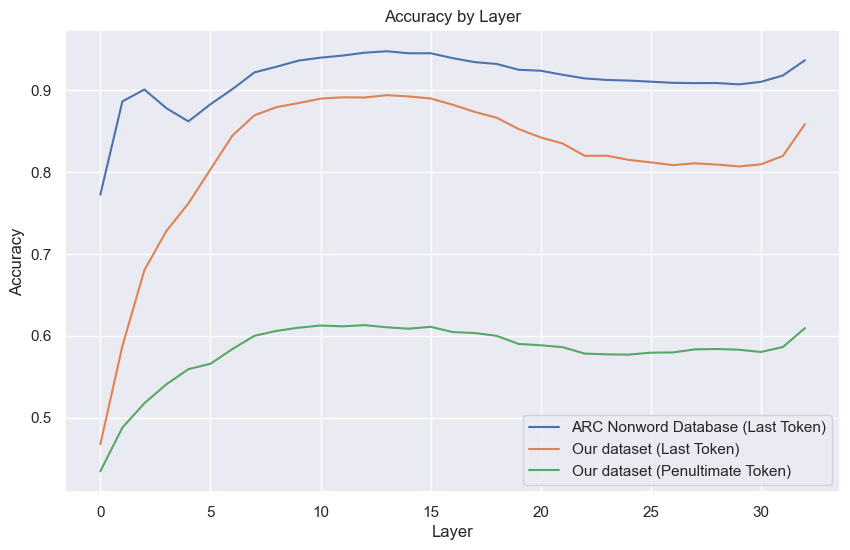}
    \caption{Accuracy comparison by layer for distinguishing words from nonwords. The ARC Nonword Database ({\textcolor{myblue}{blue}} line) exhibits the highest accuracy across layers, potentially due to its morphologically plausible nonwords. In contrast, our dataset ({\textcolor{myorange}{orange}} line, last token) achieves slightly lower accuracy, demonstrating the effectiveness of our bias-mitigated nonword generation process. The penultimate token from our dataset ({\textcolor{mygreen}{green}} line) shows significantly lower accuracy, highlighting the importance of the last token in distinguishing words from nonwords.}
    \label{fig:arc_vs_our}
\end{figure}

We further conduct additional experiments to compare the performance of our custom nonword dataset with the linguistically-motivated ARC Nonword Database~\citep{rastle2002arc}, which contains morphologically plausible nonwords designed to resemble real words to humans. Our results~(\cref{fig:arc_vs_our}) indicate a higher differentiability on the ARC dataset compared to our own dataset, which highlights that our nonword creation procedure effectively mitigates potential biases. This result strengthens our conclusions: the last token provides significant cues for distinguishing words from nonwords, particularly in the middle layers of the model after several layers of processing.

In another experiment, we evaluate the penultimate token representation against nonword tokens for words of length three or more tokens. For example, in a word like “unhappiness,” we use the inner token “h” (e.g., \textit{un-h-appiness}). Unlike the final tokens, penultimate tokens are poorly distinguishable from nonwords, achieving significantly lower accuracies. These results suggest that the model’s ability to separate words from nonwords is highly dependent on the position and completeness of token representations.

\begin{figure}[!ht]
    \centering
    \includegraphics[width=0.8\textwidth]{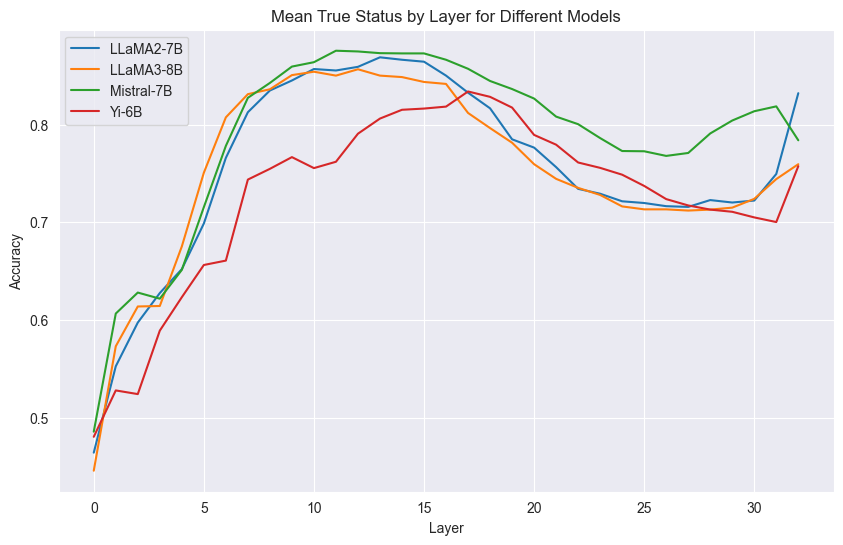}
    \caption{Mean true status accuracy by layer for distinguishing words from nonwords across multiple models (Llama2-7B, Llama3-8B, Mistral-7B, Yi-6B) using our dataset. All models exhibit similar trends: accuracy improves across initial layers, peaks in the middle layers, and declines in deeper layers. This pattern demonstrates the robustness of our findings across different model architectures and sizes.}
    \label{fig:words_vs_nonwords_multipile_models}
\end{figure}

Finally, we expand the comparison of our dataset across three additional models—Llama3-8B, Mistral-7B, and Yi-6B. Our results across these models~(\cref{fig:words_vs_nonwords_multipile_models}) mirror the trends observed with Llama2-7B, with accuracy improving as layers deepened but peaking and gradually declining in later layers. This consistency further supports our findings that the model's ability to classify words vs.~nonwords relies on nuanced patterns in internal token representations, unaffected by co-occurrence biases or morphological artifacts.

\section{Statistical Analysis of Token Aggregation in Multi-token Words}
\label{sec:appendix_token_aggregation}

\begin{figure}[htbp]
    \centering
        \begin{subfigure}{0.48\textwidth}
        \centering
        \caption{Llama2-7B Attention weights for 3-tokens words}
        \includegraphics[width=\textwidth]{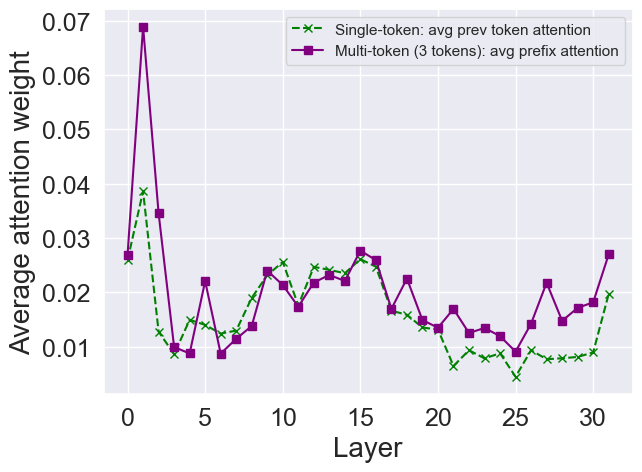}
        \label{fig:tokens_aggregation_3_tokens}
    \end{subfigure}
    \hfill
    \begin{subfigure}{0.48\textwidth}
        \centering
        \caption{Llama2-7B Attention weights for 4-tokens words}
        \includegraphics[width=\textwidth]{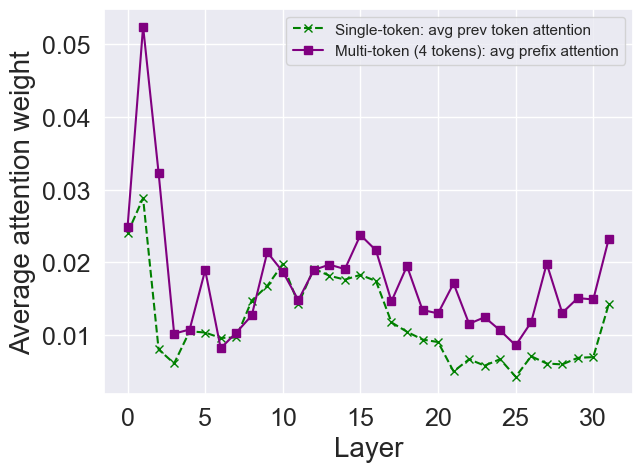}
        \label{fig:tokens_aggregation_4_tokens}
    \end{subfigure}
    \caption{Analysis for 3- and 4-token words for Llama2-7B. The higher attention pattern in layers 1-2, while lower values are observed afterwards, is consistent with our results for 2-token words~(\cref{fig:Token_Aggregation_experiment}).}
    \label{fig:tokens_aggregation_3_4_tokens}
\end{figure}

We repeat the multi-token experiments~\cref{sec:token_aggregation} for 3- and 4-token words. Our results~(\cref{fig:tokens_aggregation_3_4_tokens}) show a very similar trend to 2-token words~(\cref{fig:Token_Aggregation_experiment}).

We further present the detailed statistical analysis conducted of our experiments in~\cref{sec:token_aggregation} to 
examine \textbf{token aggregation} in multi-token words 
compared to single-token words (control group). The objective 
is to investigate whether the model exhibits significantly 
different attention patterns between these groups, indicative 
of the detokenization process.

\begin{table}[!ht]
\centering
\resizebox{0.9\textwidth}{!}{%
\begin{tabular}{cccc}
\toprule
\textbf{Layer} & \textbf{p-value (Multi $>$ Single)} & \textbf{p-value (Single $>$ Multi)} & \textbf{Significance} \\
\midrule
0  & $9.999 \times 10^{-1}$ & $5.891 \times 10^{-5}$ & Single-token $>$ Multi-token (***)
\\
1  & $0.000$                & $1.000$               & Multi-token $>$ Single-token (***)
\\
2  & $0.000$                & $1.000$               & Multi-token $>$ Single-token (***)
\\
3  & $1.000$                & $1.252 \times 10^{-273}$ & Single-token $>$ Multi-token (***)
\\
4  & $1.000$                & $0.000$               & Single-token $>$ Multi-token (***)
\\
5  & $1.000$                & $4.466 \times 10^{-242}$ & Single-token $>$ Multi-token (***)
\\
6  & $1.000$                & $0.000$               & Single-token $>$ Multi-token (***)
\\
7  & $1.000$                & $0.000$               & Single-token $>$ Multi-token (***)
\\
8  & $1.000$                & $0.000$               & Single-token $>$ Multi-token (***)
\\
9  & $1.000$                & $0.000$               & Single-token $>$ Multi-token (***)
\\
10 & $1.000$                & $0.000$               & Single-token $>$ Multi-token (***)
\\
11 & $1.000$                & $7.847 \times 10^{-87}$  & Single-token $>$ Multi-token (***)
\\
12 & $1.000$                & $6.307 \times 10^{-303}$ & Single-token $>$ Multi-token (***)
\\
13 & $1.000$                & $1.417 \times 10^{-255}$ & Single-token $>$ Multi-token (***)
\\
14 & $1.000$                & $2.397 \times 10^{-172}$ & Single-token $>$ Multi-token (***)
\\
15 & $1.000$                & $0.000$               & Single-token $>$ Multi-token (***)
\\
16 & $1.000$                & $5.857 \times 10^{-25}$ & Single-token $>$ Multi-token (***)
\\
17 & $1.000$                & $1.414 \times 10^{-152}$ & Single-token $>$ Multi-token (***)
\\
18 & $5.646 \times 10^{-5}$ & $9.999 \times 10^{-1}$ & Multi-token $>$ Single-token (***)
\\
19 & $1.000$                & $7.919 \times 10^{-161}$ & Single-token $>$ Multi-token (***)
\\
20 & $1.000$                & $2.036 \times 10^{-286}$ & Single-token $>$ Multi-token (***)
\\
21 & $1.399 \times 10^{-190}$ & $1.000$               & Multi-token $>$ Single-token (***)
\\
22 & $1.000$                & $3.647 \times 10^{-44}$ & Single-token $>$ Multi-token (***)
\\
23 & $1.000$                & $6.008 \times 10^{-13}$ & Single-token $>$ Multi-token (***)
\\
24 & $1.000$                & $1.062 \times 10^{-15}$ & Single-token $>$ Multi-token (***)
\\
25 & $5.274 \times 10^{-68}$ & $1.000$               & Multi-token $>$ Single-token (***)
\\
26 & $1.000$                & $3.670 \times 10^{-12}$ & Single-token $>$ Multi-token (***)
\\
27 & $0.000$                & $1.000$               & Multi-token $>$ Single-token (***)
\\
28 & $1.000$                & $2.446 \times 10^{-10}$ & Single-token $>$ Multi-token (***)
\\
29 & $1.438 \times 10^{-202}$ & $1.000$               & Multi-token $>$ Single-token (***)
\\
30 & $9.506 \times 10^{-99}$ & $1.000$               & Multi-token $>$ Single-token (***)
\\
31 & $6.731 \times 10^{-1}$  & $3.270 \times 10^{-1}$ & Not significant (ns)
\\
\bottomrule
\end{tabular}
}
\caption{Results of one-sided t-tests comparing attention weights between multi-token and single-token words across layers over Llama2-7B model.}
\label{tab:t-test-results}
\end{table}

To determine significance, we perform two one-sided t-tests 
per layer for the Llama2-7B case:
(1) Testing whether attention to prefix tokens in multi-token words is higher than to previous tokens in single-token words.
(2) Testing whether attention to previous tokens in single-token words is higher than to prefix tokens in multi-token words.
\Cref{tab:t-test-results} shows the p-values and significance levels for each layer. Significance levels are denoted as ns (not significant), and *** ($p < 0.001$).

The results reveal that in layers 1 and 2, attention to prefix 
tokens in multi-token words is significantly higher than in single-token words, 
suggesting the early phase of \textbf{Token Aggregation}. 
From Layers 3 to 17, attention to single-token words is higher, indicating a shift in 
attention focus from prefix attention in relation regular close tokens. 
Notably, there are intermittent increases in attention to prefix tokens at Layers 
18, 21, 25, 27, 29, and 30, possibly signaling a transfer into a prediction ensembling
phase in which the attention in general is less important \citep{artzy-schwartz-2024-attend} 
and therefore we don't see a coherent
pattern. 

In conclusion, the attention mechanism differs significantly between multi-token and single-token words, 
indicating that the detokenization process involves initial amplification of attention to 
sub-word tokens followed by a reduction as the model obtains whole-word representations.

\section{Word Retrieval for Single-Token and Multi-Token Words Across Models}
\label{sec:appendix_all_models_word_retrieval}

This section presents a detailed comparison of word retrieval performance~(\cref{sec:when_detokenization}) across several models~(Llama2-7B, Llama3-8B, Yi-6B, and Mistral-7B) for both single-token and multi-token words. The evaluation focuses on how effectively each model retrieves the original word across layers, especially in challenging cases like artificially separated single-token words, typos, and multi-token words.

\begin{figure}[!ht]
    \centering
    \includegraphics[width=1\textwidth]{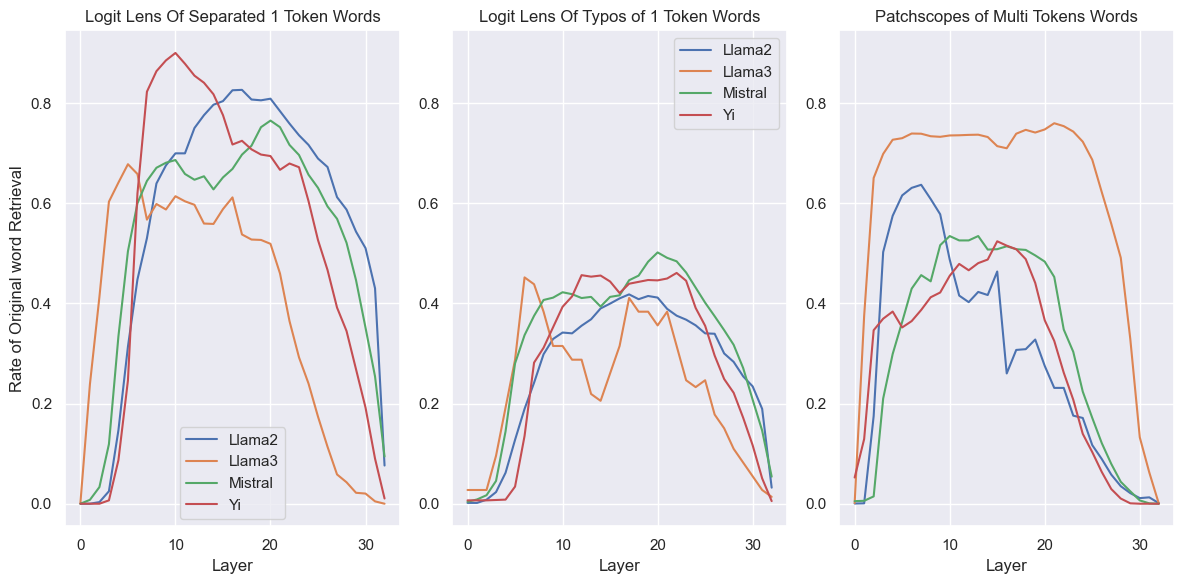}
    \caption{Layer-wise word retrieval rates for single-token and multi-token words across all models.}
    \label{fig:word_retrieval_all_models}
\end{figure}

Across all experiments~(\cref{fig:word_retrieval_all_models}), we observe a similar trend where the retrieval rate increases over the first several layers, peaks around the middle layers, and then decreases in the later layers. The main difference across models lies in the peak performance, especially in cases involving typos and multi-token words, where more advanced models such as Llama3-8B and Mistral-7B demonstrate superior performance in reconstructing the original word representations.

\subsection{Cumulative Word Retrieval for Single-Token and Multi-Token Words}
\label{sec:appendix_cummulative_word_retrieval}
\begin{figure}[tb]
    \centering
    \includegraphics[width=1\textwidth]{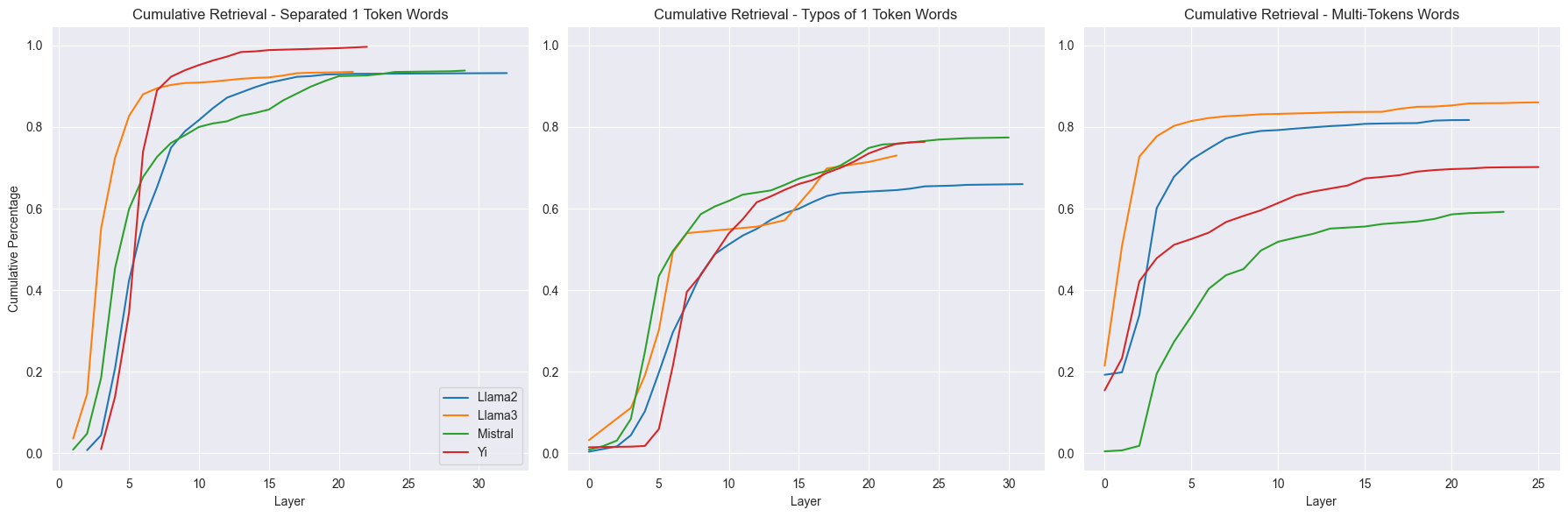}
    \caption{Cumulative word retrieval for single-token and multi-token words across all models.}
    \label{fig:word_retrieval_cummulative_all_models}
\end{figure}

\cref{fig:word_retrieval_cummulative_all_models} shows the results of cumulative word retrieval~(\cref{sec:when_detokenization}) across various models, focusing on both single-token and multi-token words. For each model, we analyze the ability of the LLM to retrieve the original word from sub-word tokens across its layers. The cumulative retrieval is calculated as the proportion of words that are successfully retrieved at each layer, with the percentage increasing as more words are recovered throughout the model’s layers.

\paragraph{Single-token words}
The cumulative word retrieval for single-token words—those that are artificially split into sub-word tokens (via typographical errors or manual splits)—shows a rapid increase in retrieval success in the early layers. For Llama2-7B, for instance, cumulative retrieval reaches 93.2\% for words split by manual intervention, and 66\% for words affected by typos by the middle layers. This pattern is observed across models, with retrieval generally peaking around layers 15-20.

\paragraph{Multi-token words}
For multi-token words, which are naturally split due to being out-of-vocabulary for the tokenizer, the cumulative retrieval process follows a similar trajectory. However, in models like Llama2-7B, the retrieval peaks earlier in the model, with a cumulative retrieval rate of 77.41\%. Other models like Llama3 and Yi show higher cumulative retrieval rates, suggesting improved efficiency in handling multi-token words, potentially due to larger model capacities and internal dictionaries.

The similarity in cumulative retrieval between single-token and multi-token words suggests that LLMs treat out-of-vocabulary words in a manner similar to sub-word-tokenized words, accessing a latent vocabulary to reconstruct full word representations.

\section{Introducing Typos for Single-Token Words}
\label{sec:appendix_introducing_typos}

In this section, we describe the process of introducing typos into single-token words to split them into multiple tokens~(\cref{sec:Single-Token_Words_Exp}). The modification applies to words longer than four characters and involves randomly performing one of three operations: substituting two characters, deleting a character, or inserting a new character. By introducing these slight variations, the word becomes unfamiliar to the tokenizer, causing it to be divided into multiple smaller tokens during tokenization. Particularly, this process results in splitting words into 2--5 tokens. \Cref{tab:typos} shows examples of the different splits.

\begin{table}[ht!]
\small
\begin{center}
\begin{tabular}{lccc}
\toprule
Description & perturbed & new tokens \\
\midrule
Substitution of two characters & devel\textbf{po}ment & {[`de', `vel', `p', `oment']}\\
Deletion of one character  & develoment & {[`de', `vel', `oment']}\\
Insertion of one character & devel\textbf{f}opment & {[`dev', `elf', `op', `ment']}\\
\bottomrule
\end{tabular}
\end{center}

\caption{\label{tab:typos} 
Examples of the different typos we consider, exemplified by perturbing the single-token word ``development''~(\cref{sec:Single-Token_Words_Exp}).} 
\end{table}

\section{Ablation Experiment on Suffix-split Words}
\label{sec:appendix_ffn_ablation}

In~\cref{sec:how_detokenization}, to test the role of FFN layers in detokenization, we conduct an intervention-based experiment measuring how ablations to FFN updates affect word retrieval.
We run this experiment on all single-token words from \wikitext that end with one of three common suffixes: \textit{``ing," ``ion,"} or \textit{``est"}.
% These suffixes are prevalent in English, making them a suitable choice for evaluating the model's ability to reconstruct original word representations.
Each word is then artificially split to two parts---the root word and the suffix. For example, the word \textit{``eating"} is split into \textit{``eat"} and \textit{``ing"}, while \textit{``connection"} is split into \textit{``connect"} and \textit{``ion"}. This ensures that processing the suffix token is necessary to reconstruct the identity of the word, and reduces possible effects of strong distributional artifacts of token co-occurrence.

Using logit lens, we identify FFN layers where the update to the residual stream can decoded as the original single-token word. We then ablate these layers by zeroing out their updates to the residual stream. As a control, we ablate the same proportion (5\%) of random FFN layers.

\begin{figure}[ht!]
    \centering
    \includegraphics[width=0.7\textwidth]{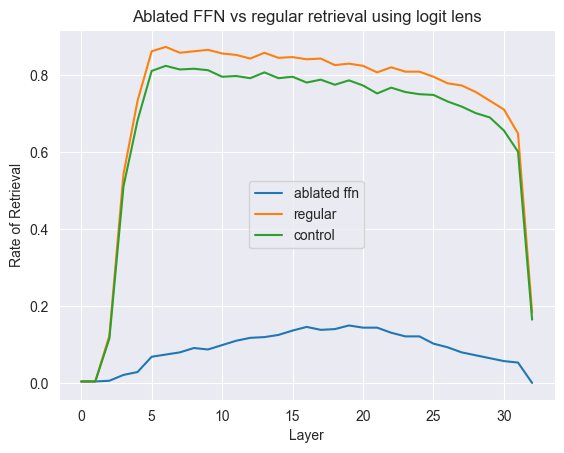}
    \caption{Comparison of retrieval rates using logit lens for suffix-split words under three conditions: ablation of identified FFN 
    layers ({\textcolor{myblue}{blue}} line), regular retrieval ({\textcolor{myorange}{orange}} line), and control with random FFN layer ablation ({\textcolor{mygreen}{green}} line). Ablation of critical 
    layers causes a sharp drop in retrieval accuracy, highlighting their importance in detokenization.}
    \label{fig:ablated_ffn_vs_regular}
\end{figure}

Our results, shown in~\cref{fig:ablated_ffn_vs_regular}, reveal that ablating the identified FFN layers lead to a sharp drop in word retrieval rates, effectively disrupting the detokenization process. In contrast, ablating random layers has little to no effect on retrieval accuracy. This indicates that the FFN layers play a critical role in reconstructing word representations rather than simply enhancing contextualization.

\section{Comparision Between Logit Lens and Cosine Similarity}
\label{sec:appendix_logit_lens_vs_cosine_similarity}

To validate our use of the logit lens, we repeat the artificial split-word experiment using cosine similarity. We use the same setting of artifical splits based on suffixes detailed above in~\cref{sec:appendix_ffn_ablation}.

% Each word is artificially split just before the suffix, ensuring that the suffix itself remains intact as an independent subword token. For example, the word \textit{``running"} is split into \textit{``runn"} and \textit{``ing"}, while \textit{``connection"} is split into \textit{``connect"} and \textit{``ion"}. This splitting process ensures that the suffix tokens~(\textit{``ing," ``ion,"} and \textit{``est"}) do not carry any distributional artifacts from the original word, meaning the retrieval of the original word could only occur through the detokenization process. This setup allows us to isolate the model's ability to reconstruct the original single-token word based solely on the hidden representation of the final token.

\begin{wrapfigure}{r}{0.48\textwidth}
\vspace{-0pt}
  \centering        
  \includegraphics[width=\textwidth]{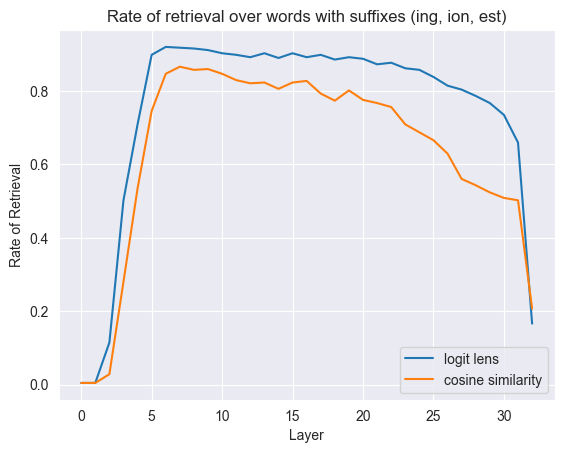}
    \caption{Comparison of retrieval accuracy for split-word experiments using either logit lens ({\textcolor{myblue}{blue}} line) and cosine similarity ({\textcolor{myorange}{orange}} line) across model layers.}
    \label{fig:logitlens_vs_cosine_similarity}
        
\vspace{0pt}
\end{wrapfigure}

For the logit lens, we adapt the standard implementation to measure similarity between the hidden representation of the final token and the input embedding space (vocabulary space), rather than the output embedding space typically used to predict the next token. This adjustment allows us to directly evaluate the alignment between the hidden states and the embeddings of the original words in the vocabulary. In parallel, cosine similarity is used to compute the similarity between the same hidden states and the embeddings of the original words in the vocabulary space.

Our results, shown in~\cref{fig:logitlens_vs_cosine_similarity}, demonstrate that both methods produce nearly identical patterns. In both cases, retrieval accuracy peaks in the middle layers, where the hidden representation of the suffix token aligns most closely with the original word. Beyond these middle layers, retrieval accuracy declines, likely reflecting the model's shift toward representing the prediction of the next token rather than maintaining the full representation of the current word.

These findings reinforce the validity of using the logit lens when adapted to the input embedding space, as a streamlined approach for analyzing model representations. This approach results in similar trends to cosine similarity while offering a more interpretable and direct framework for measuring alignment with the vocabulary space.

\section{Learning the Linear Maps \( T_{\ell, E} \) and \( T_{\ell, U} \)}
\label{sec:appending_learning_procrustes_maps}

To expand the model’s vocabulary without modifying its core parameters, we construct linear transformations that map hidden states at different layers to the model’s embedding and unembedding spaces. By learning these transformations solely from the model’s existing vocabulary, we can infer new token representations without modifying any of the model's core weights.
This section details our method for learning these mappings.

\subsection{Extracting Model Representations}
Given a pretrained language model with input embedding matrix \( E \in \mathbb{R}^{V \times d} \) and output unembedding (LM head) matrix \( U \in \mathbb{R}^{V \times d} \), where \( V \) is the vocabulary size and \( d \) is the hidden dimension, we extract the representations used for learning the transformations as follows:
\begin{itemize}
    \item \textbf{Embedding and Unembedding Matrices}: We extract the model's embedding matrix \( E \) and LM head matrix \( U \) before making any modifications to the vocabulary.
    \item \textbf{Hidden States Across Layers}: For each token \( t \) in the model’s original vocabulary, we pass \( t \) as a single-token input and record its hidden state at every layer of the model. Let \( h_{\ell}(t) \) denote the hidden state of token \( t \) at layer \( \ell \).
\end{itemize}

\subsection{Learning the Linear Mappings}
For each layer \( \ell \), we aim to learn two linear transformations:
\begin{itemize}
    \item \( T_{\ell, E} \) that maps hidden states to input embeddings.
    \item \( T_{\ell, U} \) that maps hidden states to unembedding representations.
\end{itemize}
We learn these mappings as \textbf{orthogonal Procrustes problems} \citep{schonemann1966generalized}, which seek to find the best orthogonal transformation aligning two sets of vectors. Specifically, for each layer \( \ell \), we solve:
\begin{equation}
T_{\ell, E} = \underset{T}{\arg\min} \sum_{t \in V} \| T h_{\ell}(t) - e_t \|^2, \quad \text{subject to } T^\top T = I
\end{equation}
\begin{equation}
T_{\ell, U} = \underset{T}{\arg\min} \sum_{t \in V} \| T h_{\ell}(t) - u_t \|^2, \quad \text{subject to } T^\top T = I
\end{equation}
where \( e_t \) and \( u_t \) are the embedding and unembedding vectors of token \( t \), respectively. The constraints enforce that each transformation preserves distances and does not distort the structure of the space. In our experiments, we use the Python implementation of \citet{meng2022procrustes}.

% These transformations are computed using \textbf{Singular Value Decomposition (SVD)}: given two sets of corresponding vectors \( \{h_{\ell}(t)\}_{t \in V} \) and \( \{e_t\}_{t \in V} \), we decompose their cross-covariance matrix \( C = H^\top E \) into \( C = U \Sigma V^\top \) and set \( T_{\ell, E} = UV^\top \) (and analogously for \( T_{\ell, U} \) using \( U \) instead of \( E \)).

\subsection{Normalization with RMS Scaling}
To preserve the relative magnitudes of embedding and unembedding entries, we normalize all representations using their \textbf{root mean square (RMS) norm} \citep{RMS}. Specifically:
\begin{enumerate}
    \item \textbf{Preprocessing Training Representations}: Before fitting the Procrustes transformations, we normalize all hidden states, embeddings, and unembedding vectors by dividing each vector \( x \) by its RMS norm:
    \begin{equation}
    x_{\text{norm}} = \frac{x}{\|x\|_{\text{RMS}}}, \quad \text{where } \|x\|_{\text{RMS}} = \sqrt{\frac{1}{d} \sum_{i=1}^{d} x_i^2}.
    \end{equation}
    \item \textbf{Applying the Learned Maps}: When using the trained transformations on new detokenized representations \( r \), we apply the following steps:
    \begin{enumerate}
        \item Normalize \( r \) by its RMS norm: \( r_{\text{norm}} = \frac{r}{\|r\|_{\text{RMS}}} \).
        \item Apply the learned transformation: \( \hat{e} = T_{\ell, E} r_{\text{norm}} \), \( \hat{u} = T_{\ell, U} r_{\text{norm}} \).
        \item Rescale by the mean RMS of the target space:
        \begin{equation}
        e = \hat{e} \cdot \mathbb{E}[\|e_t\|_{\text{RMS}}],
        \quad u = \hat{u} \cdot \mathbb{E}[\|u_t\|_{\text{RMS}}].
        \end{equation}
    \end{enumerate}
\end{enumerate}
This ensures that the new embeddings and unembeddings maintain a scale consistent with the original vocabulary.

\section{Efficiency Gains from Vocabulary Expansion}
\label{sec:vocab_expansion_efficency}

Beyond improving model performance on newly added words, our vocabulary expansion method directly reduces the number of tokens required to encode input text, and could lead to further potential efficiency gains in inference.

\begin{wraptable}{r}{0.55\textwidth}
\vspace{-14pt}
\resizebox{0.99\textwidth}{!}{
    \centering
    \small
    \setlength{\tabcolsep}{5pt}
    \begin{tabular}{lccc}
        \toprule
        Dataset & \# Attempted & \# New Words & Token Reduction \\
        \midrule
        \wikitext & 14.1k & 10.1k & 10.5\% \\
        \pubmed & 9.5k & 5.4k & 13.5\% \\
        \mwiki-Arabic & 4.4k & 0.7k & 14.5\% \\
        \bottomrule
    \end{tabular}
    \caption{Reduction in average sequence length when encoding text with the expanded vocabulary for Llama2-7B. \textsl{\# Attempted} is the number of multi-token words tested for expansion, while  \textsl{\# New Words} is those detected as detokenized using Patchscopes.}
    \label{tab:token_savings}
    }
    \vspace{-3pt}
\end{wraptable}

Table~\ref{tab:token_savings} summarizes the average reductions in sequence length achieved by encoding texts from each domain using the expanded vocabulary instead of the original vocabulary in the three datasets.
Using our method, we attempt to expand Llama2-7B's vocabulary with all multi-token words appearing at least $m$ times in the test set, where we use $m=1,5,50$ for \wikitext, \pubmed, and Arabic \mwiki, respectively.

We find that the token reduction rates depend on the success rate of extracting detokenized representations using Patchscopes (which in turn depend on the textual domain and language). \wikitext~achieves a 10.5\% reduction in token count, with 72.9\% of attempted words successfully converted into new vocabulary entries. \pubmed~shows a higher token savings rate (13.5\%) despite a lower success rate (58.3\%), as its vocabulary expansion targets more domain-specific multi-token terms. The largest efficiency gain is observed in Arabic \mwiki, where encoding with the expanded vocabulary reduces token count by 14.5\%, highlighting the method's potential for languages with inherently longer token sequences. However, the success rate of identifying detokenized representations in \mwiki~is lower (16.0\%), suggesting room for improvement in expanding vocabularies for morphologically rich languages, and suggesting many Arabic words are not stored in the model's inner lexicon.

These results suggest that post-hoc vocabulary expansion can significantly reduce the computational cost of inference, particularly for non-English languages and domain-specific texts, without requiring any modifications to the model's core parameters.

\end{document}